\documentclass{article}

\usepackage{amssymb}
\usepackage[english]{babel}    
\usepackage[latin1]{inputenc}   
\usepackage[dvips]{graphics}
\usepackage{subfigure}
\usepackage[pdftex]{graphicx}
\usepackage{epsfig}
\usepackage{multirow}
\usepackage{rotating}
\usepackage{picinpar}
\usepackage{color}

\usepackage{authblk}

\author[1]{JO\~{A}O BATISTA FLORINDO}

\author[2]{M\'{A}RIO DE CASTRO}

\author[3]{ODEMIR MARTINEZ BRUNO}

\affil[1]{Universidade de S\~{a}o Paulo, Instituto de F\'{i}sica de S\~{a}o Carlos, Av. Trabalhador S\~{a}o-carlense, 400\\
S\~{a}o Carlos, S\~{a}o Paulo, Brasil\\
}

\affil[2]{Universidade de S\~{a}o Paulo, Instituto de Ci\^{e}ncias Matem\'{a}ticas e de Computa\c{c}\~{a}o, Av. Trabalhador S\~{a}o-carlense, 400\\
S\~{a}o Carlos, S\~{a}o Paulo, Brasil\\
}

\affil[3]{Universidade de S\~{a}o Paulo, Instituto de F\'{i}sica de S\~{a}o Carlos, Av.  Trabalhador S\~{a}o-carlense, 400\\
S\~{a}o Carlos, S\~{a}o Paulo,Brazil\\
}
\title{Enhancing Volumetric Bouligand-Minkowski Fractal Descriptors by using Functional Data Analysis}

\begin{document}

\maketitle

\begin{abstract} 

This work proposes and study the concept of Functional Data Analysis transform, applying it to the performance improving of volumetric Bouligand-Minkowski fractal descriptors. The proposed transform consists essentially in changing the descriptors originally defined in the space of the calculus of fractal dimension into the space of coefficients used in the functional data representation of these descriptors. The transformed decriptors are used here in texture classification problems. The enhancement provided by the FDA transform is measured by comparing the transformed to the original descriptors in terms of the correctness rate in the classification of well known datasets.

\end{abstract}

%

%
%
%
%
%
%

\section{Introduction}

In recent years, the literature has presented a lot of applications of fractal theory to the solution of problems from distinct areas. As examples we may cite applications in Botany \cite{QMACG08,WSWY09,QJDPA09}, Medicine \cite{TWZ07,LC10,LSSMDB10} and Geology \cite{MG05,BDST06,DIRR07}. Particularly, in Physics, we may find applications of fractal theory in Optics \cite{SMS10,HWZ08,HDFKM07}, Materials Science \cite{CDHLAB03,W08,DAGGR09} and Electromagnetism \cite{CCWH10,SM09,VAV03}, among many other areas. Such large amount of works exploring tools from fractal theory is fully justified by an interesting observation already pointed out in \cite{M75}. This observation states that systems observed in the nature generally may be modelled by fractal measures rather than by classical formalisms.

Among the applications of fractal theory, most of them aim at using the fractal modeling in order to extract features from objects of interest according to the problem domain, like textures, contours, surfaces, etc. Such features are then provided as input data, for example, to methods for segmentation, classification and description of objects. A classical example of such fractal feature is the fractal dimension.

As in the most of cases the simple use of fractal dimension is still not sufficient to well represent the complexity of an object or scenario from the real world, the literature developed techniques for the extraction of a set of features based on the fractal dimension. Examples of such approaches are Multifractal theory \cite{H01,LRAJ08,LGS00}, Multiscale Fractal Dimension (MFD) \cite{MCSM02,CC00} and Fractal Descriptors \cite{BPFC08,BCB09,PPFVOB05,FCB10}. 

Here, we are focused on fractal descriptors approach. Several authors, like in \cite{PPFVOB05,BPFC08,BCB09,FCB10}, obtained interesting results in different applications of fractal descriptors technique to texture and shape analysis, mainly in the description of natural objects. Particularly, here we are focused on an approach developed in \cite{BCB09} which uses the volumetric Bouligand-Minkowski fractal dimension to generate a set of descriptors. Such descriptors obtained a high performance in an application to a task of plant leaves classification based on texture. 

Nevertheless, an important drawback of fractal descriptors technique, particularly that based on Bouligand-Minkowski, is that the curve formed by the set of descriptors present a high correlation, that is, each descriptor is strongly dependent on each other. This correlation does their performance decrease drastically in problems of classification and segmentation with a high number of samples and classes. In such situations, volumetric Bouligand-Minkowski descriptors have severe limitations.

Aimimg at enhancing Bouligand-Minkowski descriptors, preserving the reliability of the results, this work proposes the development and use of Functional Data Analysis (FDA) transform concept. Functional Data Analysis is a powerful statistical tool developed in \cite{RS02}. It represents an alternative to the traditional multivariate approach and deals with complex data as being a simple analytical function: the functional data. FDA approach presents certain advantages in this kind of application, like the easy handling of data in nonlinear domains (as the case in Bouligand-Minkowski descriptors) and the intuitive notion of functional operations, like derivatives and smoothing, employed in the definition of fractal descriptors.

Up our knowledge, Florindo et al. \cite{FCB10} is the first work to apply the FDA approach to fractal descriptors. In that work, functional data representation is used for reducing the dimensionality of the descriptors set in shape recognition problems. Here, we propose a different paradigm for FDA use, by defining the concept of FDA transform. The FDA transform is defined as the operation which changes the original data set (in this case, descriptors) space into the space of coefficients of functional data. The transform still presents two variants: the first uses the coefficient directly, the second performs a second algebraic transform, described in \cite{RDCV05}.

The relevance of the FDA transform is verified in experiments of classification of two well known datasets, that is, Brodatz \cite{B66} and OuTex \cite{OMPVKH02}. The results are compared in terms of classification correctness rate. It was considered two variants of the FDA transform and it was compared through three classifiers very well known in the literature: Linear Discriminant Analysis (LDA), K-Nearest Neighbors (KNN) and Bayesian \cite{DH00,B07,F90}.

This work is divided into seven sections, including this introduction. The following explains the concepts of fractal theory, fractal dimension and fractal descriptors. The third introduces the Functional Data Analysis theory and definitions. The fourth shows the proposed method. The fifth describes the experiments. The sixth section shows the results and the last section concludes the work.

\section{Fractal Analysis}

The literature shows a lot of applications of fractal geometry involving the characterization of natural objects and scenarios. Examples of such applications may be found in \cite{SMS10,CDHLAB03,CCWH10,WSWY09,BDST06,HWZ08,DAGGR09,VAV03}. Most of these works use the fractal dimension as a metric for describe the object. This strategy is justified by the fact that fractal dimension measures the complexity of a structure. Physically, the complexity corresponds to the irregularity or to the spatial occupation. These properties are tightly related to constitution aspects which allow the identification of such objects.

An important drawback of using only fractal dimension is that it is a unique global value and is not capable of extract information about intricate details of a structure. With the aim of exploring fully the potential of fractal theory, the literature shows the development of techniques which provide not only a unique value but a set of values capable of describing in a richer way an object, based on the fractal theory. Among these techniques, we have the Multifractal \cite{H01,LRAJ08,LGS00}, the Multiscale Fractal Dimension \cite{MCSM02,CC00} and the Fractal Descriptors \cite{BPFC08,BCB09,PPFVOB05,FCB10}. 

Multifractal theory replaces the fractal dimension analysis by the concept of fractal spectrum, capable of modeling objects which cannot be represented by a single fractal measure. Multifractal demonstrates to be an interesting tool to capture the different power-law scaling present in a system \cite{H01,LRAJ08,LGS00}.

The literature still shows an alternative technique for the modeling of objects with fractal theory. This approach is the Multiscale Fractal Dimension (MFD) \cite{MCSM02,CC00}. In MFD approach, instead of simply calculate the fractal dimension from interest objects, a set of features is extracted from the derivative of the whole power-law curve used to provide the fractal dimension.

\begin{figure}[!htb] 
					 \centering
           \mbox{\subfigure[]{\epsfig{figure=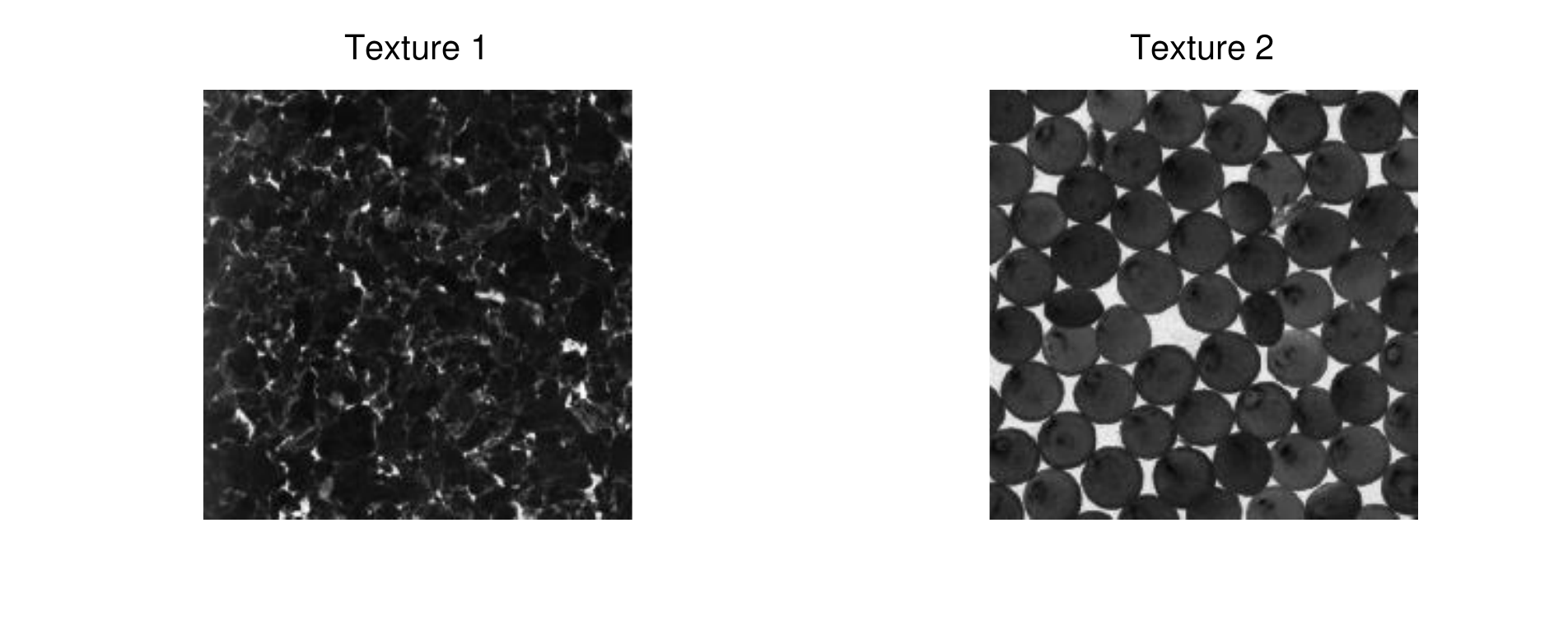,width=\textwidth}}}
           \mbox{\subfigure[]{\epsfig{figure=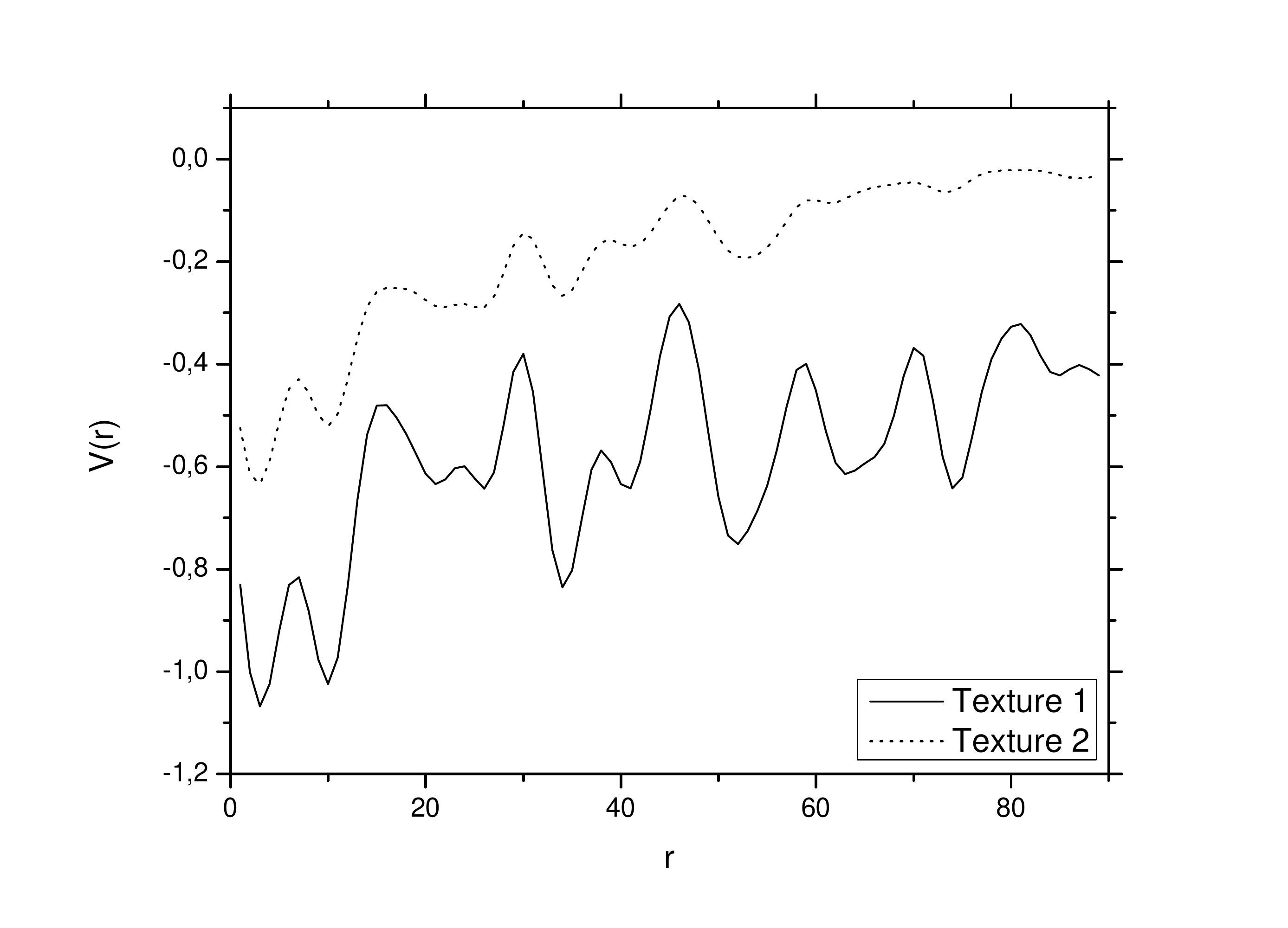,width=\textwidth}}}           			 
           \caption{Two textures with the same fractal dimension present fractal descriptors totally different. (a) Original textures (both with fractal dimension 2.618). (b) Fractal descriptors from the same textures.}
           \label{fig:MFD1}                                  
   \end{figure}

An extension of MFD are the fractal descriptors \cite{BPFC08,BCB09,PPFVOB05,FCB10}. In this case, we extract features (descriptors) from an object through the calculus of the fractal dimension taking the object under different observation scales. These descriptors are used to compose a feature vector that could be mean as a ``signature'' to characterize the object. Particularly, fractal descriptors demonstrate to be an efficient tool for the discrimination of natural textures like that analyzed in the present work. The Figure \ref{fig:MFD1} illustrates the discrimination power of fractal descriptors $D(k)$ by showing two distinct textures whose fractal dimensions are identical but the curve of fractal descriptors is visually distinct.

The following sections describe in more details the aspects involved in fractal descriptors technique, starting from the fractal dimension definition.

\subsection{Fractal Dimension}
\label{sec:dimension}

Fractal dimension is a real positive number constituting the main measure extracted from a fractal object. There is no absolute definition for the concept of fractal dimension. The most used and classical one is the Hausdorff-Besicovitch dimension.

Hausdorff-Besicovitch dimension $\dim_{H}(F)$ is a concept derived from the measure theory and is defined over a set $F \subset \Re^{n}$ as
\begin{equation}
	\dim_{H}(F) = \{s\} | \inf \left\{ s:H^{s}(F)=0 \right\} = \sup \left\{ H^{s}(F)=\infty \right\},
\end{equation}
where $H^{s}(F)$ is the $s-$dimensional Hausdorff-Besicovitch measure, defined by
\begin{equation}
	H^{s}(F) = \lim_{\epsilon \rightarrow 0}{H_{\epsilon}^{s}(F)},
\end{equation}
where
\begin{equation}
	H_{\epsilon}^{s}(F) = \inf\left\{ \sum_{i=1}^{\infty}{|U_{i}|^{s}:{U_{i} \mbox{ is an $\epsilon$-cover of F}}} \right\}.
\end{equation}
In above equations, $||$ expresses for the diameter in $\Re^{n}$, that is, \linebreak $|U| = \sup{|x-y|:x,y \in U}$.

In many situations, the calculus of Hausdorff-Besicovitch dimension is very complex and even impracticable. In such cases, we can calculate it by generalizing the concept of classical Euclidean dimension \cite{M75}. In this way, we obtain the following expression
\begin{equation}\label{eq:HB}
	\dim_{H}(F) = \lim_{\epsilon\rightarrow 0}\frac{\log(N(\epsilon))}{\log(\frac{1}{\epsilon})},
\end{equation}
where $N(\epsilon)$ is the minimum number of objects with linear size $\epsilon$ needed to cover $F$ \cite{F86}.

Most of different definitions of fractal dimension are based on a generalization of Equation \ref{eq:HB}, expressed through
\begin{equation}
\label{eq:dimension}
	D = \lim_{\delta \rightarrow 0}\frac{\log(M_{\delta}(S))}{\log(\delta)},
\end{equation}
where $M$ is a set measure depending on the specific fractal dimension method and $\delta$ is the scale parameter. As example of fractal dimensions defined from the previous expression we can cite the box-counting, the packing dimension, the Renyi dimension, etc \cite{F86}.

Particularly, here we are focused on the Bouligand-Minkowski fractal dimension \cite{F86}. As the Hausdorff-Besicovitch dimension, the Bouligand-Minkowski dimension also is based on a topological measure, in this case, the Bouligand-Minkowski measure $meas_{M}$ calculated through
\begin{equation}
	meas_{M}(F,S,\tau) = \lim_{r \rightarrow 0}\frac{V(\partial F \oplus rS)}{r^{n-\tau}},
\end{equation}
where $F$ is the object (set) of interest, $S$ is a structuring element with radius $r$ and $V$ is the volume of the dilation between $S$ and the boudary $\partial C$ of $C$. The Bouligand-Minkowski dimension itself is given by
\begin{equation}
	\dim_{M}(F,S) = \inf \left\{ \tau,meas_{M}(F,S,\tau) = 0 \right\}.
\end{equation}

For an application to discrete objects represented in a digital image, the calculus is significantly simplified through the use of neighborhood techniques. In this way, the above expression becomes
\begin{equation}
	\dim_{M}(F) = \lim_{\epsilon \rightarrow 0}\left( N-\frac{\log C(F \oplus S_{\epsilon})}{\log(\epsilon)} \right),
\end{equation}
in which $S$ is a disk with diameter $\epsilon$ (also called dilation radius), $C$ is the number of points pertaining to the dilation region $F \oplus S_{\epsilon}$ and $N$ is the topological dimension of the space in which $F$ is immersed.

\subsection{Multiscale Fractal Dimension}

Although fractal dimension is an important measure, it is insufficient for a good representation of complex systems which present different fractal dimension depending on the observation scale taken into account. In order to provide a richer fractal-based information from an object, the literature shows the Multiscale Fractal Dimension (MFD) \cite{MCSM02,CC00}.

MFD consists in the application of a multiscale transform to the fractal dimension. The multiscale transform of a signal $u(t)$ is the function $U(b,a)$, where $b$ is directly associated with $t$ and $a$ is the scale variable. Essentially, the multiscale is performed through three approaches: scale-space, time-frequency and time-scale. In the following, we describe the approach used in MFD, e.g., scale-space. More details are found in \cite{CC00}.

Scale-space is a particular case of multiscale transform. It is based on the derivative of the signal followed by a convolution with a smoothing gaussian filter \cite{W84}:
\[
	\{(b,a)|a,b \in \Re, a > 0, b \in \{U^1(t,a)\}_zc\},
\]
where $._zc$ expresses the zero-crossings $.$ and $U^1(t,a)$ represents the convolution of the original signal $u(t)$ with the first derivative of the gaussian $g^1_a$, that is:
\[
	U^1(t,a) = u(t) * g_a^1(t).
\]

In \cite{MCSM02}, the MFD is obtained from the Bouligand-Minkowski fractal dimension in the following manner:
\begin{equation}
	MFD_a  = 2 - \frac{d(log(A(r)))}{d(\log(r))} * g_a,
\end{equation}
where $A(r)$ is the dilation area for each dilation radius $r$. In MFD technique, some characteristics of MFD curve, like maximum, minimum and area below the curve graph, are extracted to compose a feature vector for the analyzed object.

\subsection{Fractal Descriptors}

Fractal descriptors \cite{BPFC08,BCB09,PPFVOB05,FCB10} are an extension of MFD concept where a feature vector is extracted from the fractal dimension calculated over a whole interval of scales.

Generally speaking, fractal descriptors are obtained from the function $u$: 
\[
	u:\log(\epsilon) \rightarrow \log(M(\epsilon)),
\]
where $M$ is a measure depending on the fractal dimension estimation method and $\epsilon$ is the scale parameter. 

\begin{figure}[!htb] 
					 \centering
           \mbox{\subfigure[]{\epsfig{figure=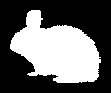,width=0.2\textwidth}}
           			 \subfigure[]{\epsfig{figure=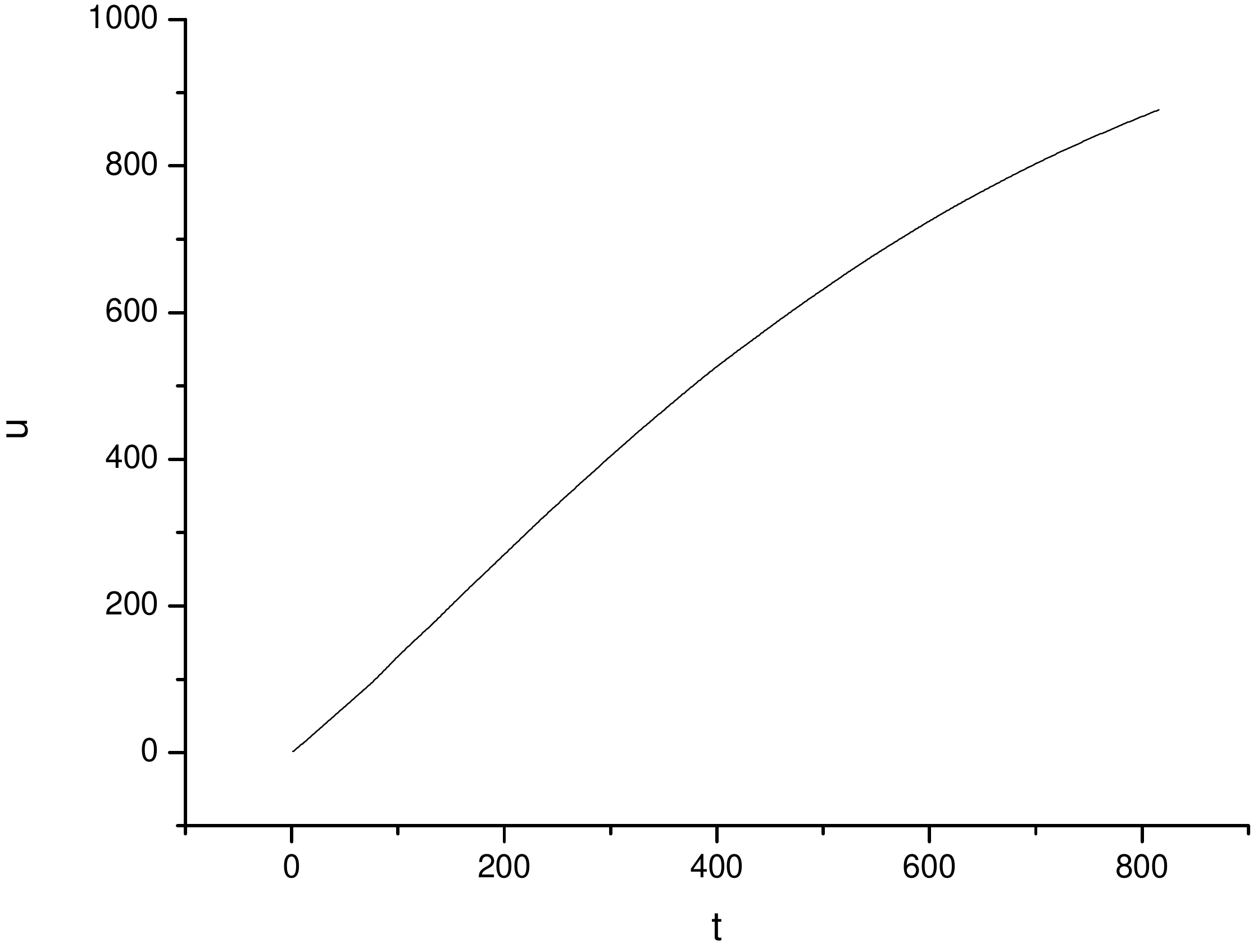,width=0.4\textwidth}}
           			 \subfigure[]{\epsfig{figure=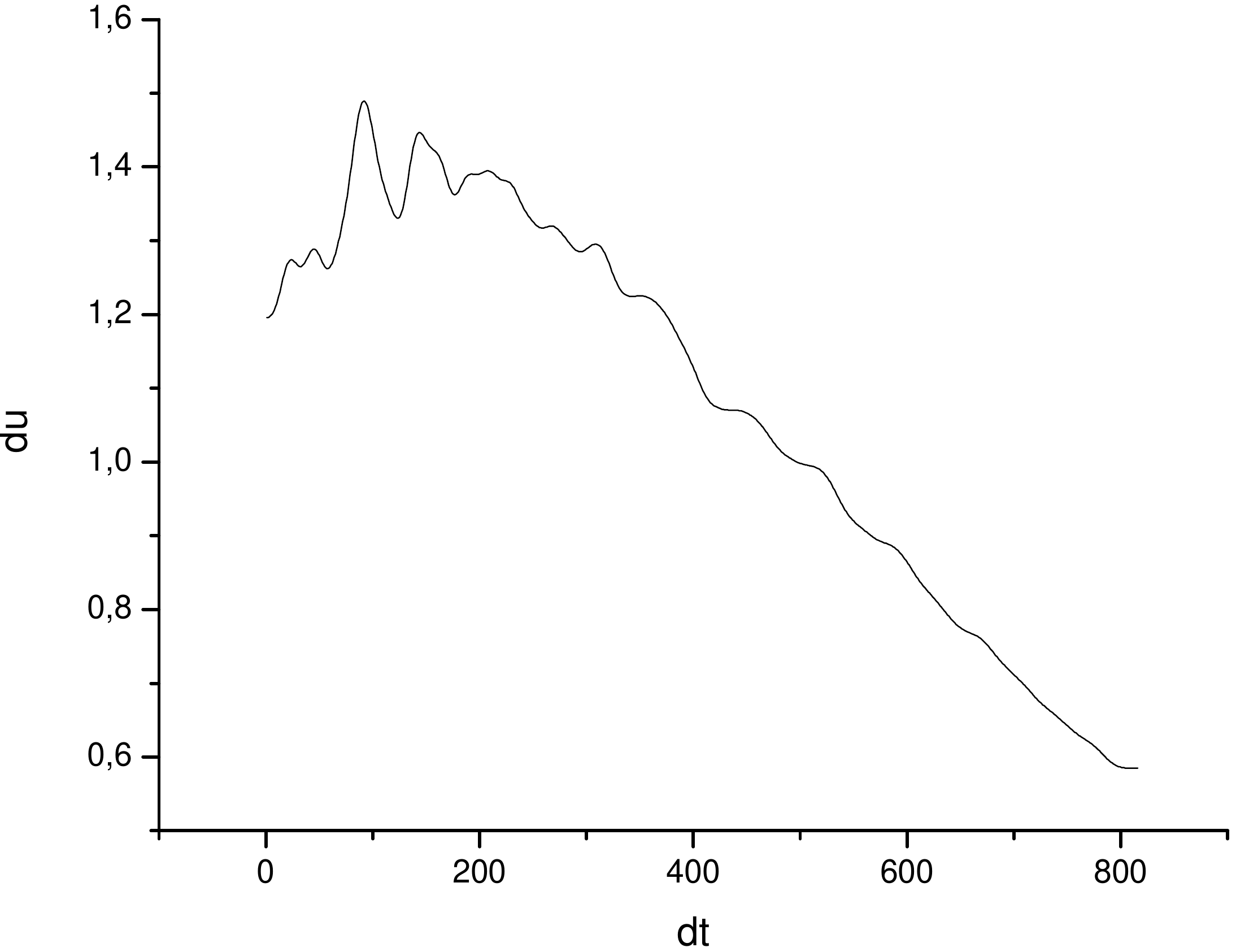,width=0.4\textwidth}}}           			 
           \caption{Fractal descriptors curve. (a) Object analyzed. (b) Curve $u(t)$. (c) Descriptors curve after Fourier derivative, followed by gaussian convolution.}
           \label{fig:MFD3}                                  
   \end{figure}

The function $u$ must be used directly, as in \cite{BCB09}, or may be summited to a particular transform. For instance, in \cite{BPFC08}, the descriptors $\mathfrak{D}$ are extracted from the Fourier derivative of $u$:
\[
	\mathfrak{D} = \frac{du}{dt} = \mathfrak{T}^{-1}(D(f)U(f)),
\]
where $t$ is equivalent to $\log(\epsilon)$, $U$ is the Fourier transform of $u$ and $D$ is Fourier derivative:
\[
	D(f) = j2\pi u,
\]
where $j$ is the imaginary number. In order to attenuate noises inherent to the derivative operation, one may still apply a convolution with a gaussian filter embedded in the Fourier derivative, as employed in \cite{FCB10}. Thus, the above expression becomes:
\[
	\mathfrak{D} = \frac{du}{dt} = \mathfrak{T}^{-1}(D(f)U(f)G^1_a),
\]
where $G^1_a$ is the derivative of the gaussian $g_a$ in the Fourier domain. The Figure \ref{fig:MFD3} shows the aspect of descriptors curve of an object.

In \cite{PPFVOB05}, the descriptors are obtained from the Fourier derivative, followed by a Principal Component Analysis (PCA) transform, aiming at reduce correlation among descriptors. In this way, a more reliable and consistent set of descriptors are provided to characterize plant leaf shapes analyzed in that work.

Here, we propose the application of Functional Data Analysis, described in the following, as a transform to $u$, in order to generate more robust and precise fractal descriptors.

\subsection{Volumetric Bouligand-Minkowski Fractal Descriptors}

In this work we focus on a specific fractal descriptors approach developed in \cite{BCB09} called Volumetric Bouligand-Minkowski Fractal Descriptors (VBFD). The main idea is the calculus of Bouligand-Minkowski fractal dimension of a 3D surface taken under a range of observation scales. These descriptors are employed to describe texture images, that is, analysis of images based on spatial and color arrangement of pixels.

\begin{figure}[!htb] 
					 \centering
           \mbox{\subfigure[]{\epsfig{figure=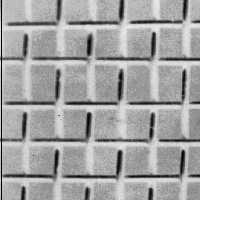,width=0.3\textwidth}}
           			 \subfigure[]{\epsfig{figure=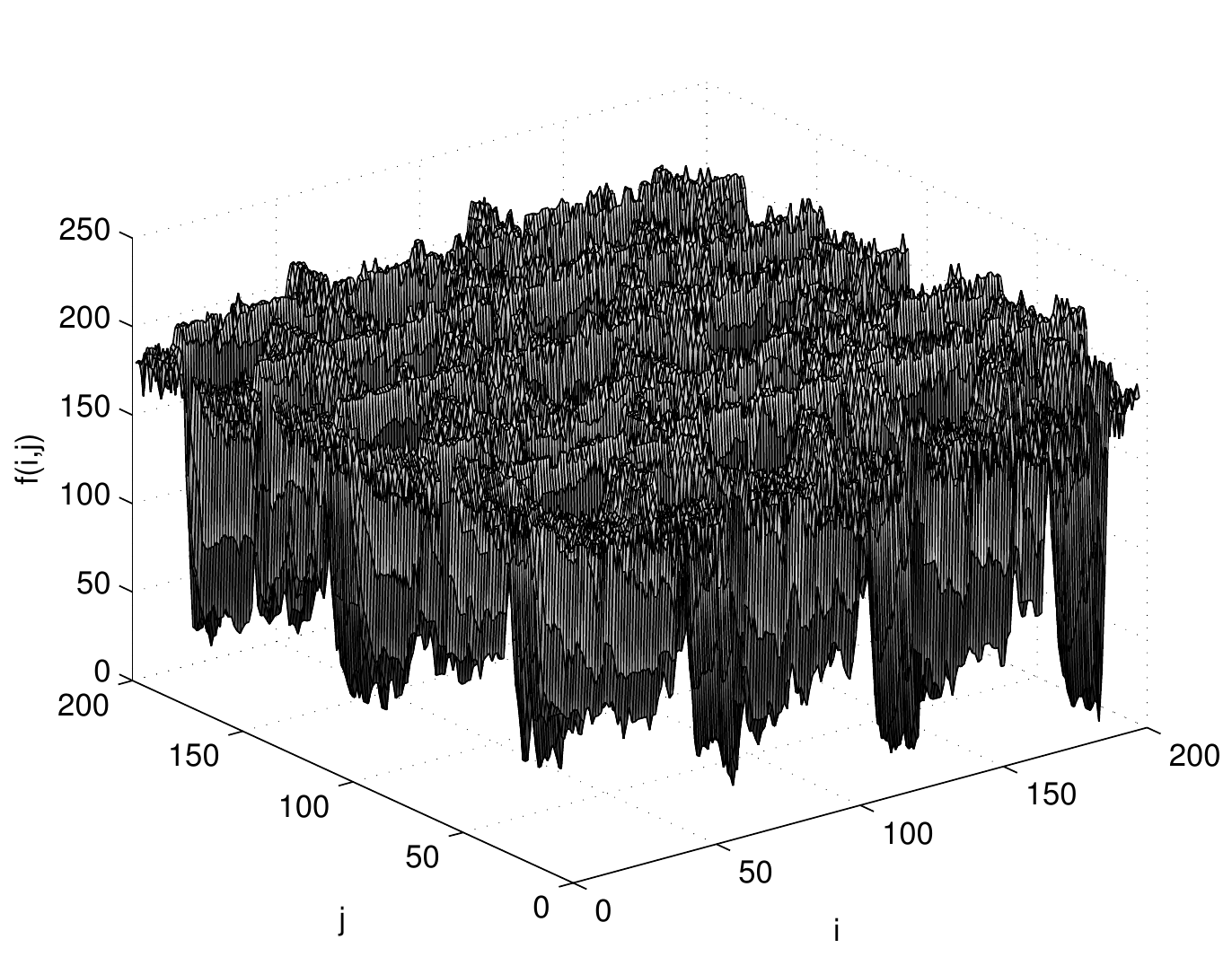,width=0.6\textwidth}}}           			 
           \caption{Texture mapped onto a surface. (a) Original texture. (b) Threedimensional surface.}
           \label{fig:surf}                                  
   \end{figure}

In the first step, we map the intensity image $Img \in [1:M] \times [1:N] \rightarrow \Re$ onto a threedimensional surface
\begin{equation}
	Surf = \{i,j,f(i,j)|(i,j) \in [1:M] \times [1:N]\},
\end{equation}
such that
\begin{equation}
	f(i,j) = \{1,2,...,max\_gray\}|f = Img(i,j),
\end{equation}
where $max\_gray$ is the maximum pixel intensity. This transform is illustrated in the Figure \ref{fig:surf}.

\begin{figure}[!htb] 
					 \centering
           \mbox{\subfigure[]{\epsfig{figure=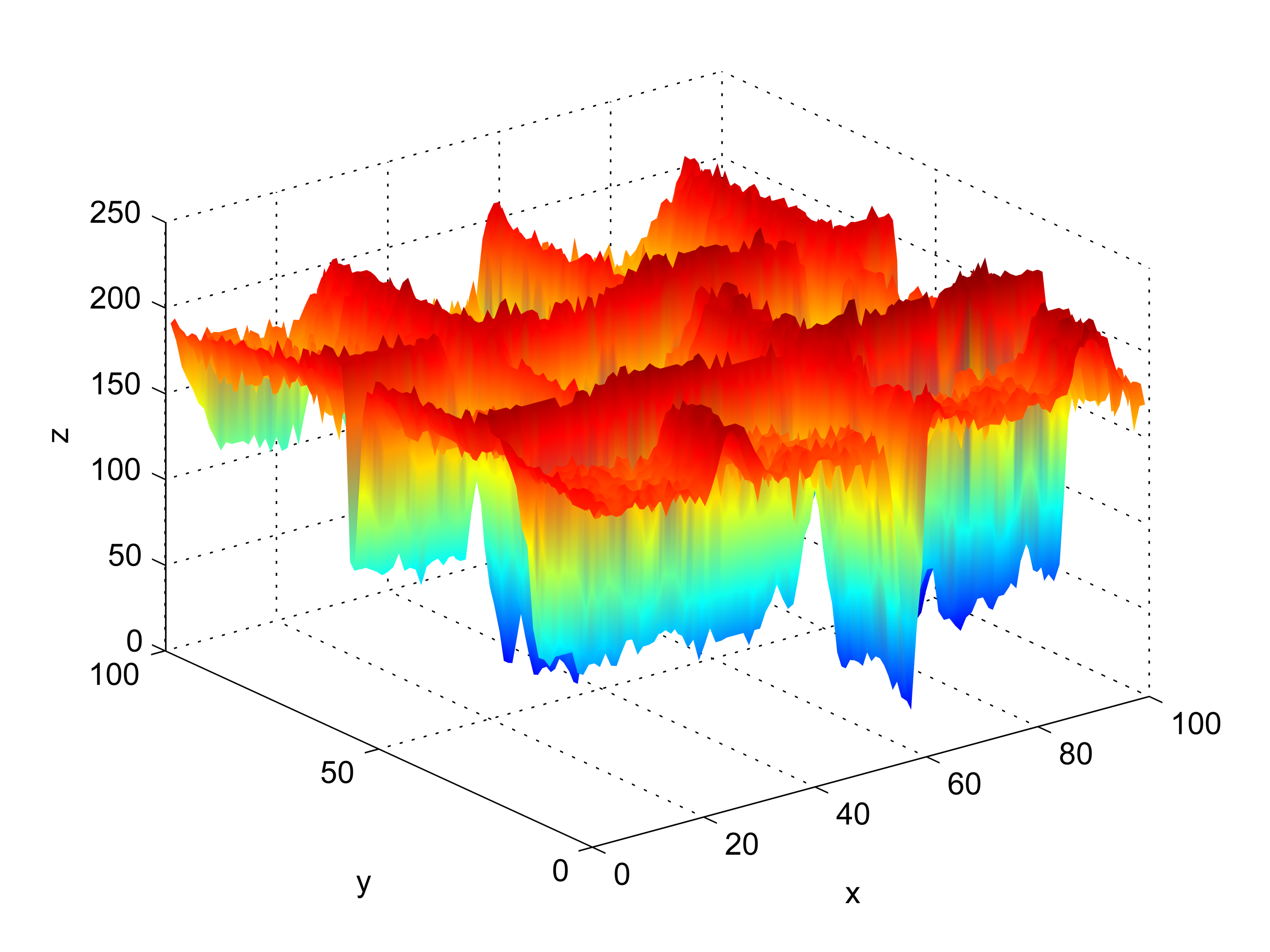,width=0.5\textwidth}}
           			 \subfigure[]{\epsfig{figure=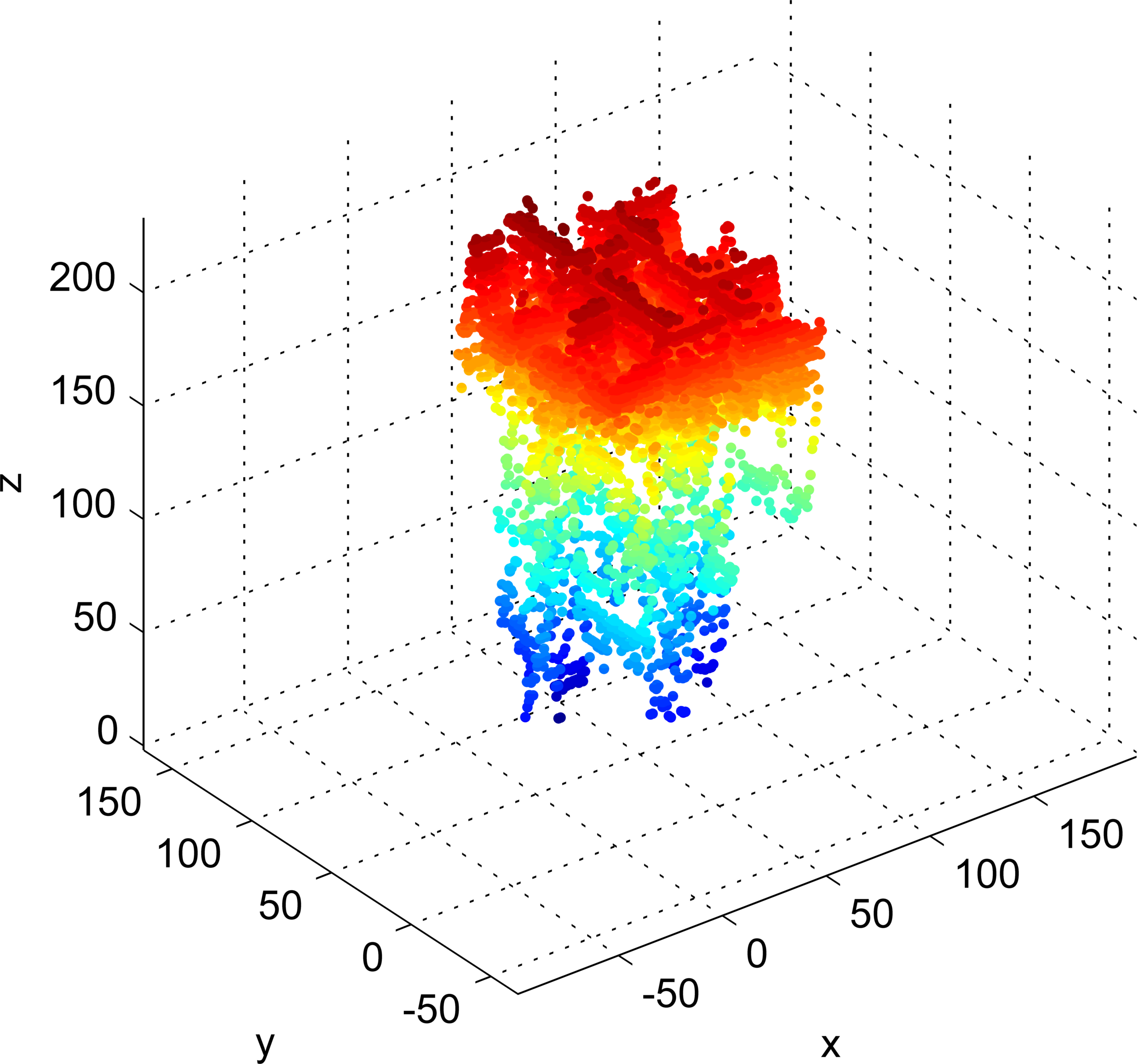,width=0.5\textwidth}}}
           \mbox{\subfigure[]{\epsfig{figure=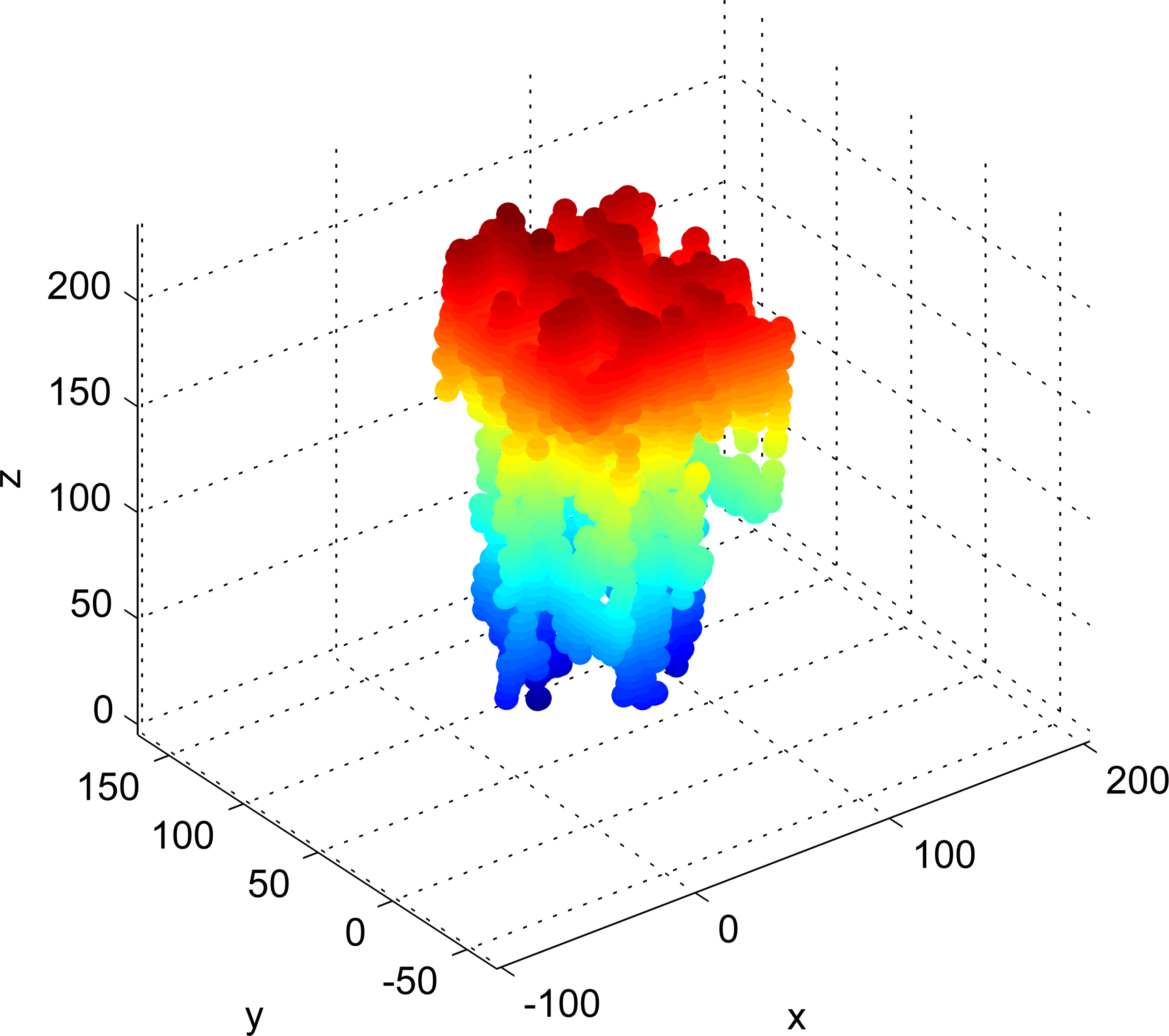,width=0.5\textwidth}}
           			 \subfigure[]{\epsfig{figure=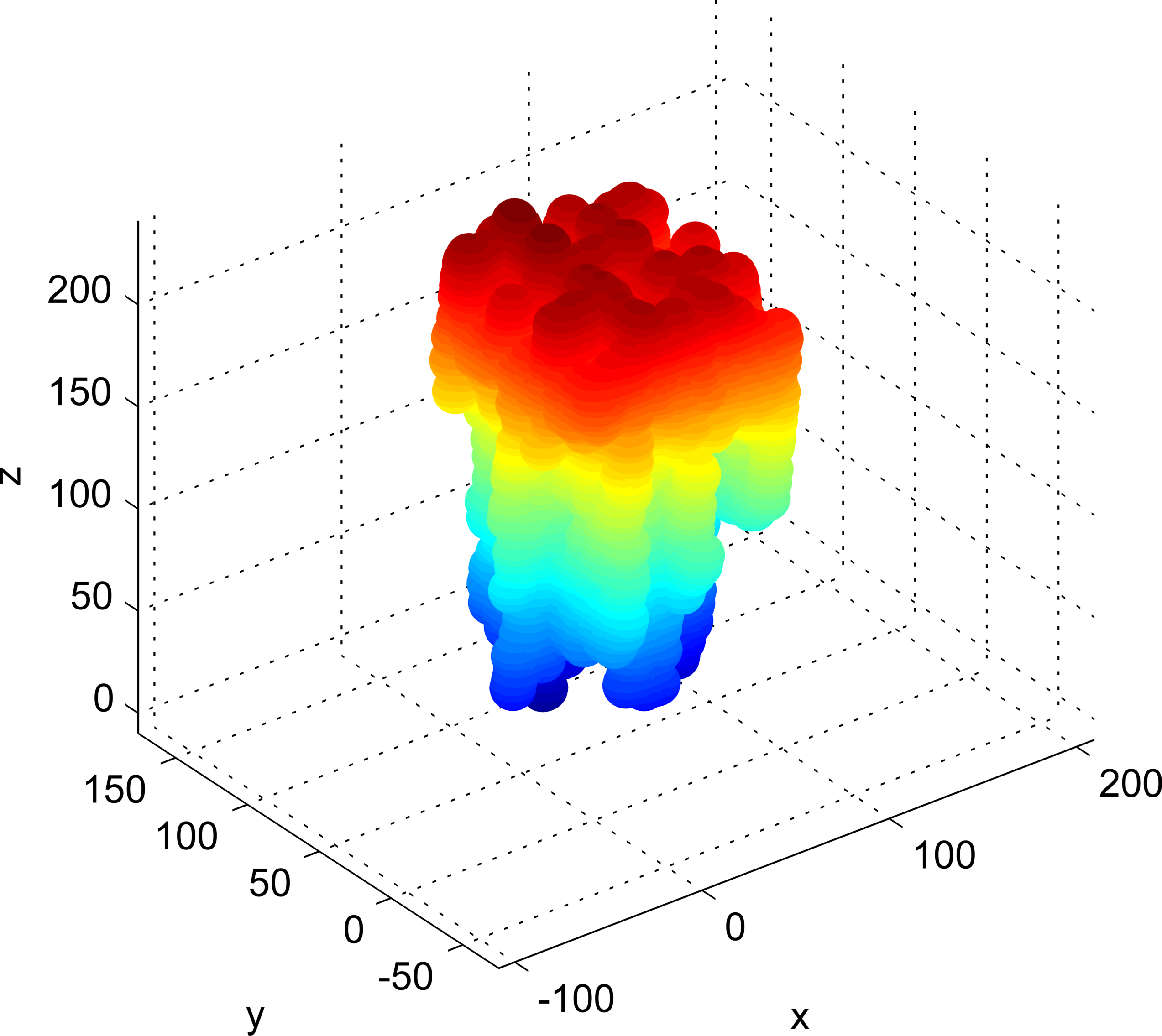,width=0.5\textwidth}}}           			 
           \caption{Dilated surfaces with different radii. (a) Points from original surface. (b) Radius 2. (c) Radius 5. (d) Radius 10.}
           \label{fig:dilat}                                  
   \end{figure}

In the following, each point of the surface is dilated by a sphere with variable radius $r$, like illustrated in the Figure \ref{fig:dilat}. Finally, we analyze the dilation volume $V(r)$, that is the number of points inside the structure composed by the dilation with each radius $r$. $V(r)$ also corresponds to the number of points with a distance at most $r$ from the object. Thus, the exact Euclidean Distance Transform (EDT) \cite{FCTB08} becomes an efficient tool for this calculus.

In 3D space, EDT is defined as the distance of each point in the space to a subset of it. In our case, this subset is the surface and the EDT for each point outside $Surf$ is given by
\begin{equation}
	EDT(p) = \min\{d(p,q)|q \in Surf^{c}\},
\end{equation}
in which $d$ is the Euclidean distance.

Exact EDT is characterized by the fact that distances present discrete values $E$
\begin{equation}
	E = {0,1,\sqrt{2},...,l,...},
\end{equation}
where
\begin{equation}
	l \in D = \{d | d = (i^2+j^2)^{1/2};i,j \in \mathbb{N}\}
\end{equation}

The dilation volume is provided by
\begin{equation}
	V(r) = \sum_{i=1}^{r}{Q(i)},
\end{equation}
where
\begin{equation}
	Q(r) = {(x,y,z)|g_k(P) - [g_r(P) \cap \cup_{i=0}^{r-1}g_i(P)]},
\end{equation}
such that
\begin{equation}
	 g_r(P) = \left\{ \begin{array}{l}
	 (x,y,z)|[(x-P_x)^2 + (y-P_y)^2 \\
	 + (z-P_z)^2]^{1/2} = E(r);x,y,z \in N
	 \end{array} \right\},
\end{equation}
where
\begin{equation}
	P = {(x,y,z) | f(x,y,z) \in Surf}
\end{equation}

The Bouligand-Minkowski fractal dimension $FD$ of the surface is simply given through
\begin{equation}
	FD = 3 - \alpha
\end{equation}
where $\alpha$ is the slope of a straight line fit to the curve $\log(V(r)\times\log(r))$. The technique in \cite{BCB09} went beyond the simple fractal dimension calculus and uses all values $\log(V(r))$ as descriptors for texture: the VBFD descriptors. Notice that $\log(V(r))$ is directly related to the Bouligand-Minkowski dimension for maximum radius $r$. Moreover, each radius corresponds to an observation scale, from further (greater radius) to closer (smaller radius). 

\begin{figure}[!htb] 
					 \centering
           \mbox{\subfigure[]{\epsfig{figure=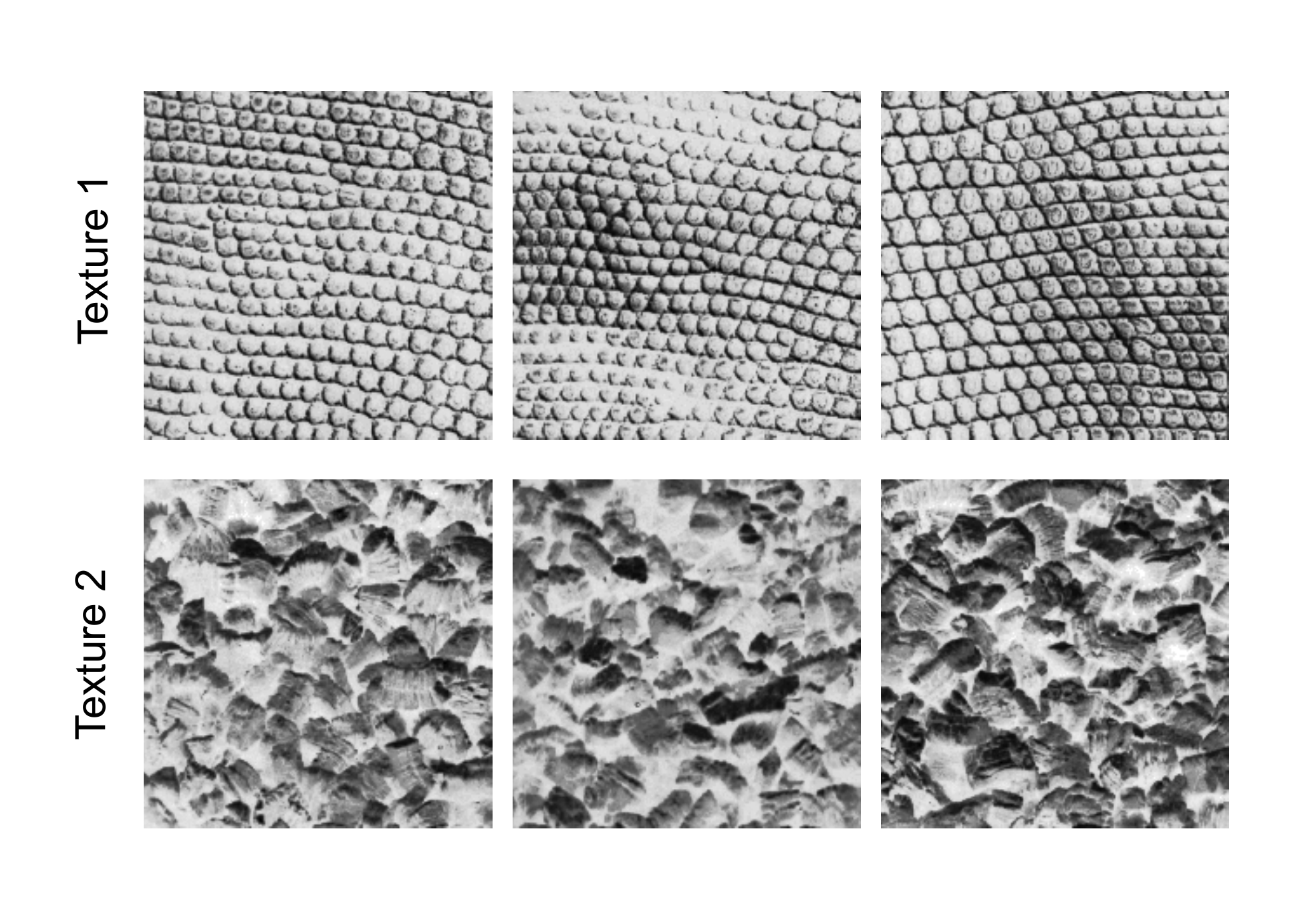,width=0.9\textwidth}}}
           \mbox{\subfigure[]{\epsfig{figure=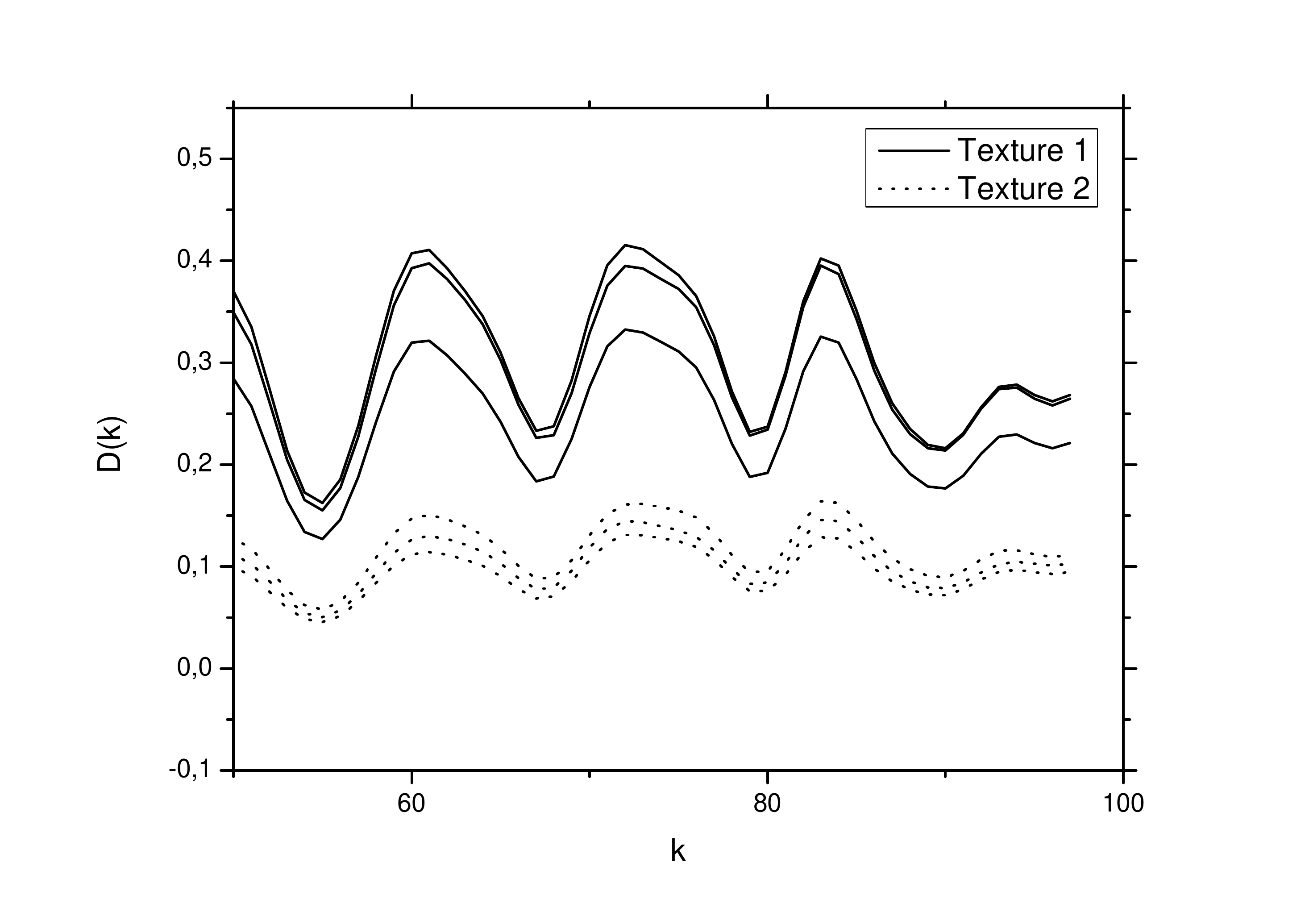,width=0.9\textwidth}}}           			 
           \caption{Discrimination power of fractal descriptors. (a) Textures from two different materials. (b) Fractal descriptors from each texture in a single graph.}
           \label{fig:MFD2}                                  
   \end{figure}

In that work, they employed VBFD in the discrimination of plant leaves images, achieving excellent results. The Figure \ref{fig:MFD2} shows the capability of MFD curve in discriminating two textures from different materials.   
   
Here, we propose the use of Functional Data Analysis \cite{RS02} in order to enhance the performance of volumetric Bouligand-Minkowski descriptors.

\section{Functional Data Analysis}

Functional Data Analysis (FDA) \cite{RS02,RS05} is a statistical tool alternative to multivariate analysis. While in multivariate statistic, we are interested in relations among observations of variables, in FDA, each observation of a set of variables is handled as a unique analytical function. Thus we extract measures from those functions, like derivatives, curvature, etc. In FDA terminology, each observation is called ``data'' and the function is called ``functional data''. 

FDA has found applications in Economy, Biology, Meteorology, etc. like synthesized in \cite{RS05}. The functional representation has some noticeable advantages in practical applications. For example, we simplify the global analysis of an observation with missing values, irregular sampling domain or noise. Besides, the analytical representation allows the application of operations like derivatives of different orders, curvatures, integrals, etc. These operations turn possible, for example, a more accurate analysis of the variability level of the data among other important characteristics.

The main result of FDA theory attests that any observation of statistical variables with analytically smooth behavior may be perfectly represented by some analytical function. In practice, this function is not exactly known but we may obtain efficient approximations. The most common strategy for the obtainment of such approximations is the interpolation of the data through  specific basis functions, like Fourier, wavelets, polynomials. Here we choose the use of B-splines functions, due to their flexibility in data representation. This process is also known as basis function development.

\begin{figure}[!htb] 
					 \centering
           \includegraphics[width=\textwidth]{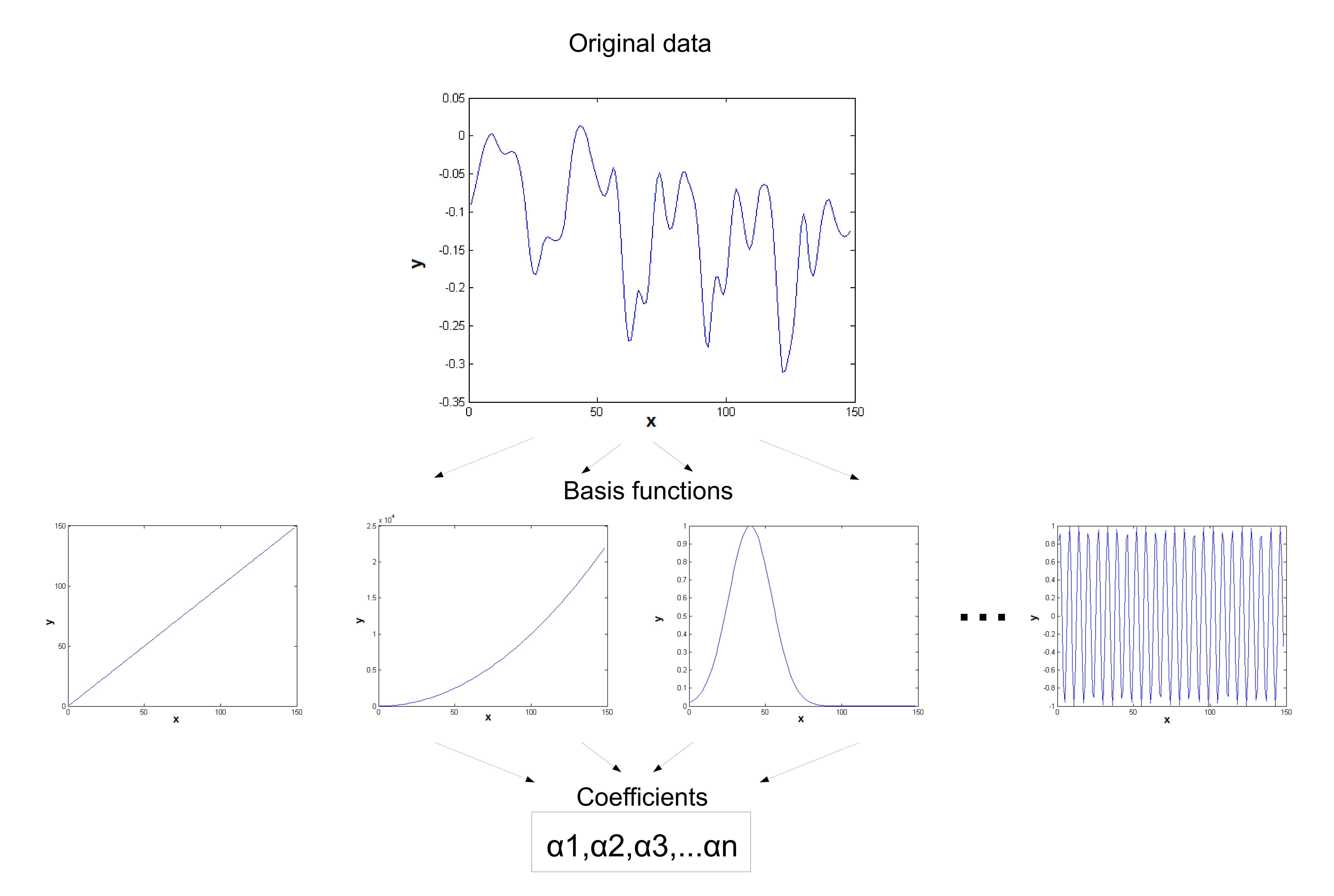}          			 
           \caption{Functional data representation. From above, the curve of an original measure, the basis functions and the coefficients used to represent computationally  the functional data.}
           \label{fig:FDA2}                                  
   \end{figure}

For the computational statistical analysis we must represent the analytical function numerically in some manner. Then, the coefficients of basis functions in the interpolation are used as the effective functional representation and statistical metrics, like mean and variance, are calculated from these coefficients. Here we also use a more complex representation in which the coefficients are previously summited to an algebraic transform. The Figure \ref{fig:FDA2} illustrates FDA representation.

\subsection{Mathematical Foundations}

For computational purposes, the data which must be analyzed consists in the observation of discrete pairs of variables $u = (x_{j}^{i},y_{j}^{i})_{1 \leq j \leq m^{i},1 \leq i \leq n}$, in which $x_{j}^{i} \in V$, $y_{j}^{i} \in \Re$ are the pair values in the variable $i$ and $m^{i}$ is the number of observed pairs for each variable. For example, an observation may be the measure of temperature along a day and each pair may correspond to the hour in which the temperature is measured and the respective temperature value.

The fact is that in this hypothetical example, the temperature cannot be perfectly modelled by any continous function, capable of predicting exactly the temperature for any day and hour. In order to apply the FDA concept to such data, we need to use a very important result from FDA theory which affirms that a data capable of being analyzed through FDA may be represented by an analytical function if we take into consideration a tolerable approximation noise. The result demonstrates that we can find an analytical function $u^{i}$ such that
\begin{equation}
 y_{j}^{i} = u^{i}(x_{j}^{i})+\epsilon_{j}^{i},
\end{equation}
where $\epsilon_{j}^{i}$ measures the noise inherent to the acquisition process.

Although the function $u$ is not known explicitly, some classical function interpolation techniques have been applied to the observation pairs yielding an approximation of $u$. Generally, this approximation method is based on the development of $u$ into functions basis. This technique consists in the projection of approximating function $u^{i'}$ in a subspace with $q$ linearly independent basis functions $(\phi_{i})_{1 \leq i \leq q}$. In this way, the approximating function is represented by $u^{i'} = \sum_{j=1}^{q}{\alpha_{j}(u^{i'})\phi_{j}}$. The values of $\alpha_{j}(u^{i'})$ are called the projection coefficients and are calculated as those which minimizes
\begin{equation}
 \sum_{j=1}^{m^{i}}{\{y_{j}^{i}-\sum_{k=1}^{q}{\alpha_{k}(u^{i'})\phi_{k}(x_{j}^{i})}\}^{2}},i=1,...,n.
\end{equation} 
Particularly, in this work, the functions basis choosen were the B-splines, which already demonstrated good results in the data here analyzed as shown in \cite{FCB10}. In the following, we describe briefly the B-splines concept.

\subsection{B-splines}

B-spline is a particular kind of spline function. A spline is a real function composed by piecewise polynomial functions. 

More formally speaking, a spline may be defined as
\begin{equation}
	s:[a,b] \rightarrow \Re.
\end{equation}
The interval $[a,b]$ is divided into $n$ ``knots'' $k$ as
\begin{equation}
	a = k_1 < k_2 < ... < k_n = b.
\end{equation}
In each subinterval $[k_{i-1},k_i]$, the spline $s$ is given by the polynomial $P_i$. The order of the spline corresponds to the highest order of polynomials. Each polynomial $P_i$ is called a basis of the spline function.

A B-spline is a particular category of splines characterized by minimum support (number of points where the function has value different of zero). Each B-spline basis $B_{i,j}$ is defined through
\begin{equation}
	B_{i,0}(t) = \left\{ \begin{array}{l}
		1 \mbox{, if } t_i \leq t < t_{i+1} \mbox{ and }t_i < t_{i+1}\\
		0 \mbox{, otherwise.}
	\end{array}\right.
\end{equation}
\begin{equation}
	B_{i,j}(t) = \frac{t-t_i}{t_{i+j}-t_i}B_{i,j-1}(t) + \frac{t_{i+j+1}-t}{t_{i+j+1}-t_{i+1}}B_{i+1,j-1}(t)
\end{equation}

Finally, the B-spline curve is given by
\begin{equation}
	B(t) = \sum_{i = 0}^{n}{L_{i}B_{i,p}(t)},
\end{equation}
where $p$ is the degree of the basis and $L_i$ corresponds to the knots.

\section{Proposed Method}
\label{sec:method}

Beyond its importance as a statistical analysis tool, FDA has demonstrated to be an efficient technique to extract relevant information from a large data set. For example, in \cite{H05}, a large amount of data respect to the water quality is collected in a specific local. Thus, the FDA approximating function is obtained from each curve of observed values and coefficients $\alpha_{j}(u^{i'})$ are used to extract important characteristics from data.

In \cite{FCB10}, FDA is used to reduce the dimensionality and extract utile information from fractal descriptors used in a task of shape analysis. In that case, instead of using directly $\alpha_{j}(u^{i'})$ coefficients, it is employed a transform of that coefficients which takes into account the contribution of the function basis space used.

This transform is performed by the canonical transform matrix $\Phi$
\begin{equation}
	\Phi(k,l) = <\phi_{k},\phi_{l}>,
\end{equation}
where $\phi$ are the basis functions. Besides, to simplify the notation, we use $\alpha(u) = (\alpha_{1}(u),...,\alpha_{q}(u))$, corresponding to the set of coefficients from the $q$ basis functions. Thus, the transformed coefficients $\beta(u)$ are given through
\begin{equation}
	\beta(u) = S\alpha(u),
\end{equation} 
where $S$ is the result from the matrix $\Phi$ decomposed by the Choleski method \cite{RDCV05}, that is, $S$ is a unique lower triangular decomposition matrix of $\Phi$, such that $\Phi = SS^*$, where $S^*$ is the conjugate transpose of $S$.

\begin{figure}[!htb] 
					 \centering
           \mbox{\subfigure[]{\epsfig{figure=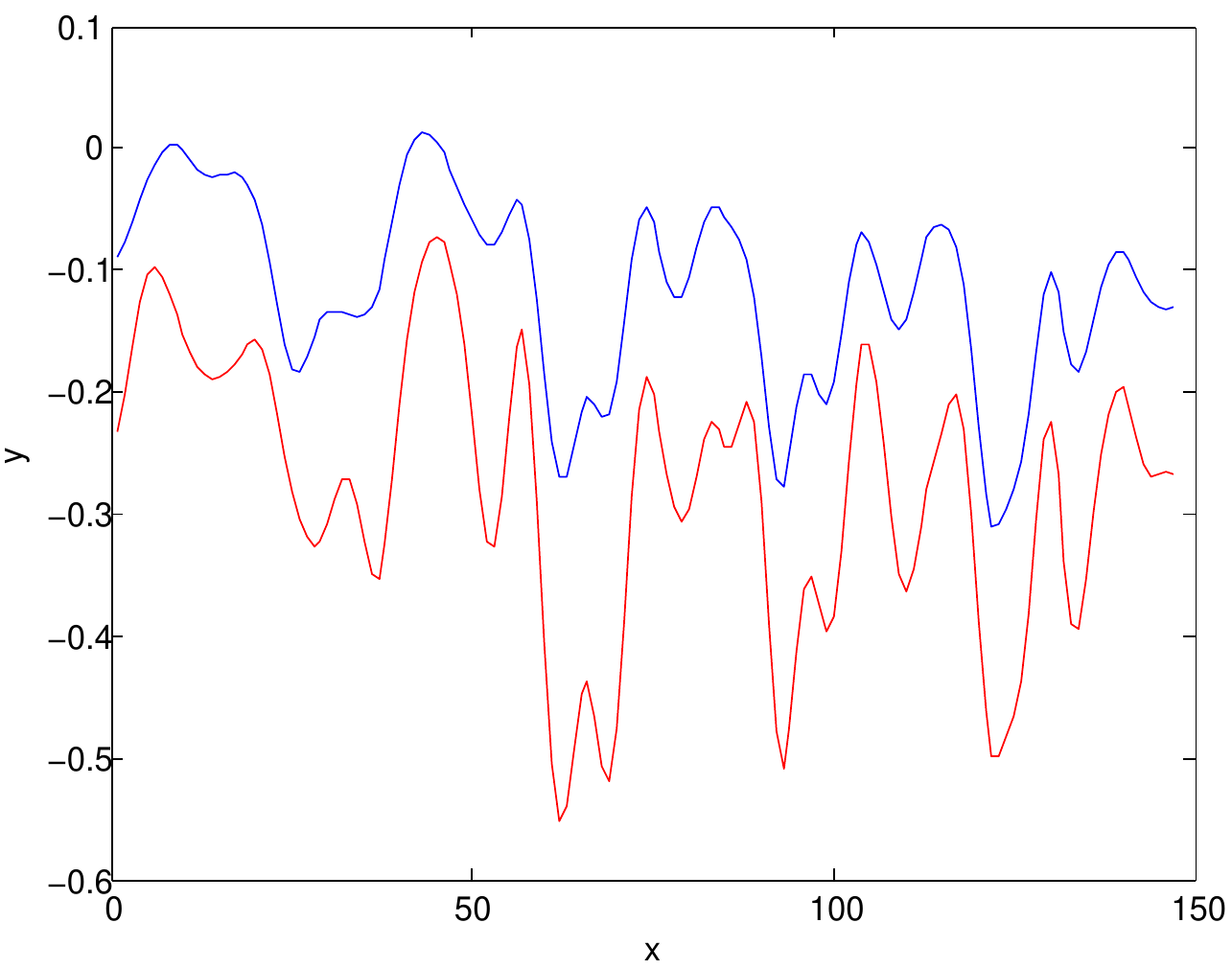,width=0.33\textwidth}}
           			 \subfigure[]{\epsfig{figure=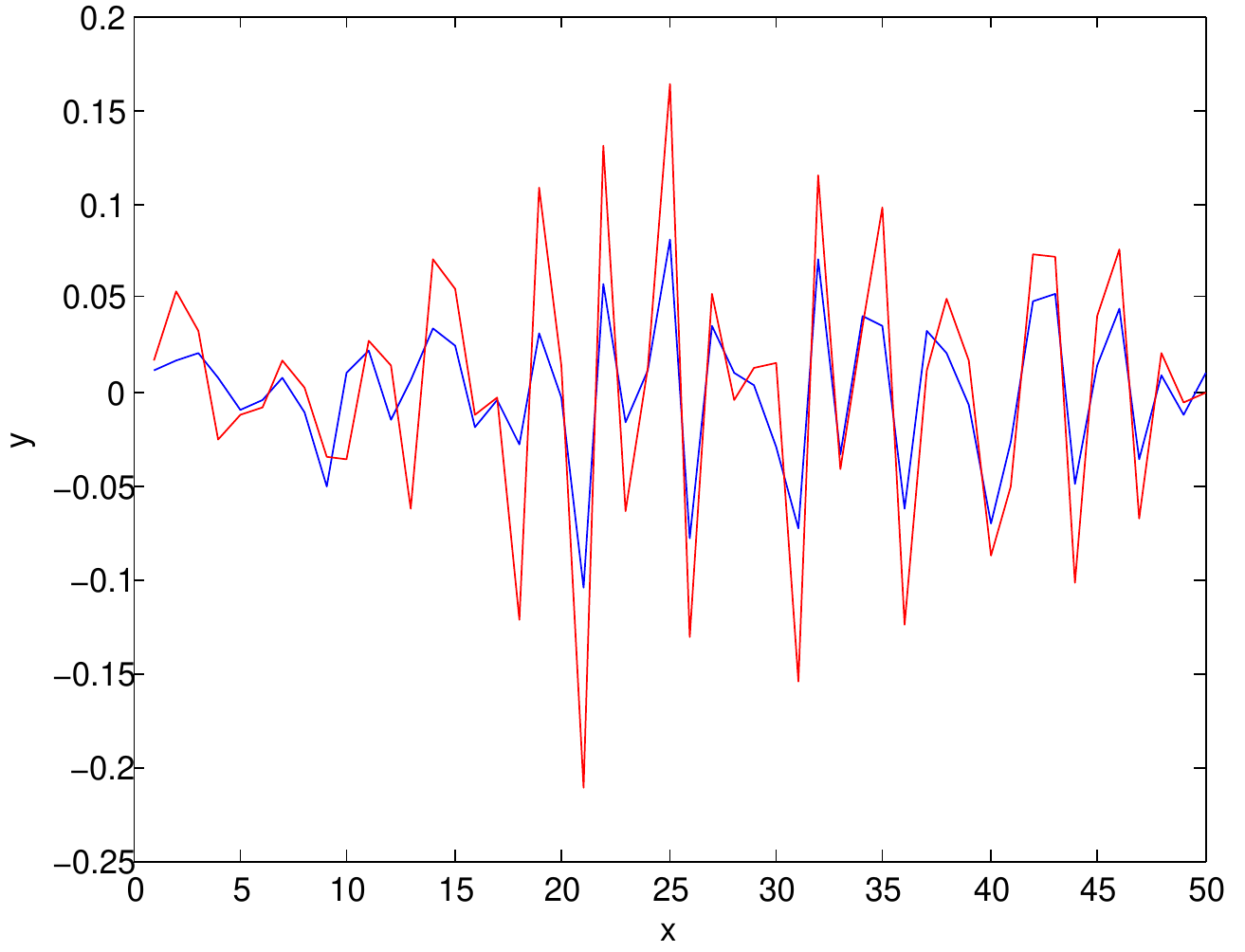,width=0.33\textwidth}}
           			 \subfigure[]{\epsfig{figure=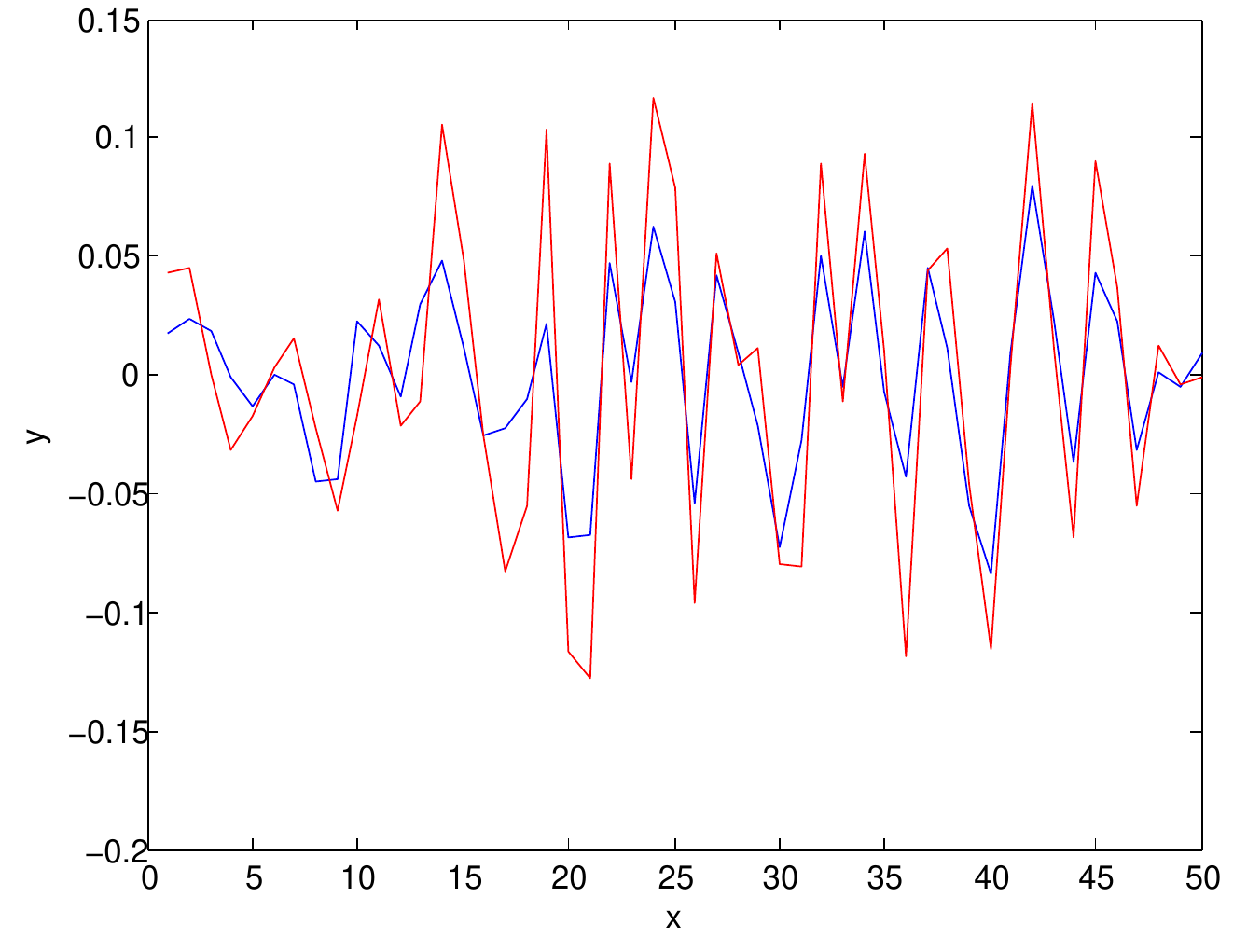,width=0.33\textwidth}}}           			 
           \caption{Discriminative property of FDA. (a) Original data curve. (b) FDA coefficients. (c) Transformed coefficients.}
           \label{fig:FDA3}                                  
   \end{figure}

The Figure \ref{fig:FDA3} shows the discriminative power of FDA. We observe a data represented in two curves with similar visual aspect and the discrimination of FDA coefficients without and with transform.

The present work extends the application in \cite{FCB10} to the analysis of volumetric Bouligand-Minkowski descriptors, applied to texture classification. The motivation for this application comes from the fact that volumetric Bouligand-Minkowski descriptors corresponds to a typical case of data whose FDA representation is interesting, according to \cite{RS02,RS05}. In fact, the descriptors present a global smooth aspect, being therefore analytical. Besides, they are extracted from a nonlinear space (log-log curve) and then are provided by a domain irregularly spaced. Another motivation is the fact that fractal descriptors may involve a derivative operation which becomes more intuitive by the handling of descriptors as a function and not only as a simple set of non related values.

Unlike the situation in \cite{FCB10}, the objective of using FDA here is not the simple dimensionality reduction, even because volumetric descriptors are easily treated by traditional classification methods. A problem with volumetric descriptors is that, although they allow for the achievemet of good results, they present a high level of correlation, that is, the original descriptors have a high dependence among themselves. This fact implies in difficulties for discrimination tasks envolving a large number of samples and classes. Our purpose is to evidentiate patterns in the global structure of descriptors which turn possible the enhancement of the discrimination power of original volumetric descriptors.

\begin{figure}[!htb] 
					 \centering
           \epsfig{figure=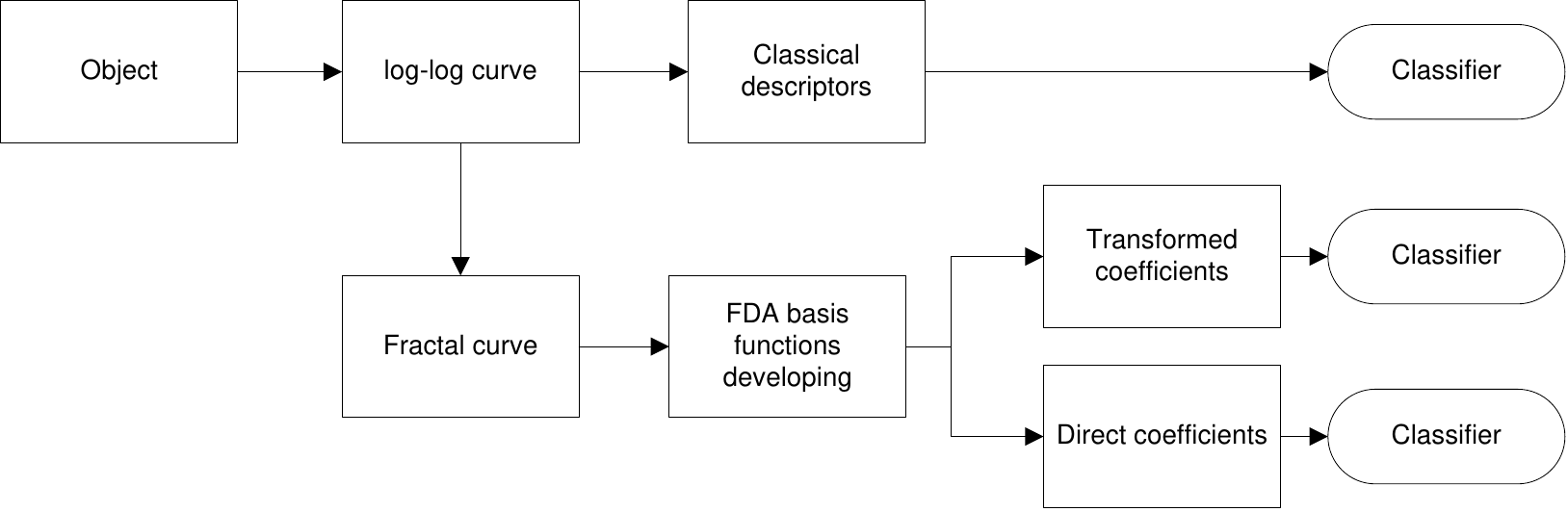,width=\textwidth}          			 
           \caption{Functional data representation. A diagram illustrating the steps in the FDA transform applied to fractal descriptors.}
           \label{fig:fda}                                  
   \end{figure}

For this goal, we propose the use of direct coefficients $\alpha(u)$ or transformed coefficients $\beta(u)$ replacing conventional fractal descriptors in a classification method. We call this representation form as FDA transform. In fact, we have a typical transform, in which the data is mapped from the log-log space of volumetric Bouligand-Minkowski descriptors onto the space of coefficients of functional data. The Figure \ref{fig:fda} illustrates the FDA transform steps.

\section{Experiments}

The performance of the FDA transform on volumetric Bouligand-Minkowski fractal descriptors is tested in an application to the classification of textures from two different datasets.

The first is the classical Brodatz texture dataset \cite{B66}, composed by 111 classes, each one with 10 samples (images) corresponding to photographies of real world textures. The second analyzed dataset is the also well known Outex dataset \cite{OMPVKH02}, composed by 68 classes with 20 images in each class. 
\begin{figure}[!htb]
	\centering
		\includegraphics[width=\textwidth]{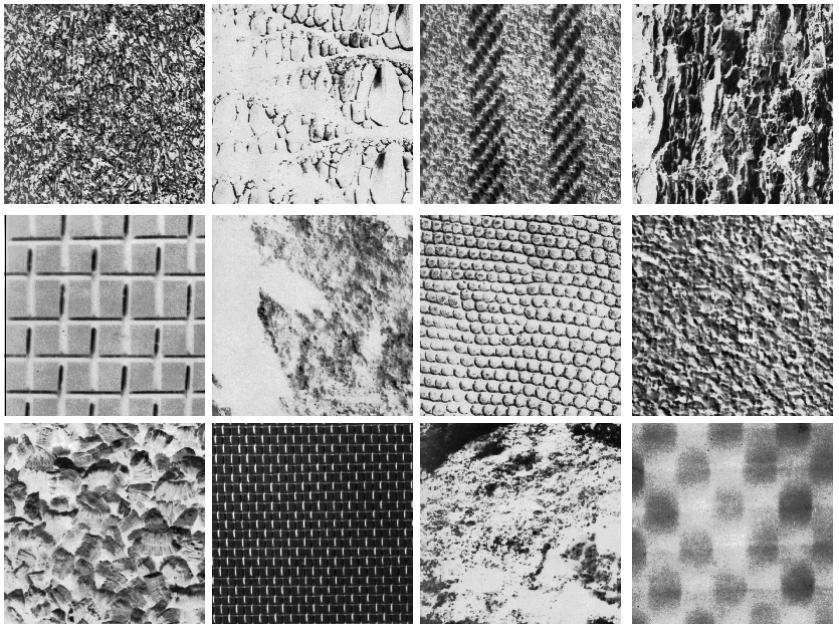}
	\caption{Some examples of images from the Brodatz dataset.}
	\label{fig:brodatz}
\end{figure}
\begin{figure}[!htb]
	\centering
		\includegraphics[width=\textwidth]{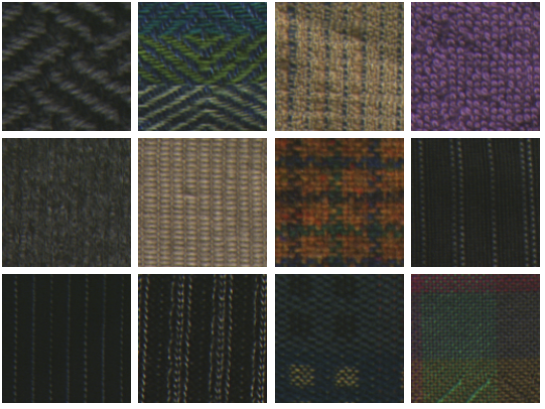}
	\caption{Some examples of images from the OuTex dataset.}
	\label{fig:outex}
\end{figure}

The steps in the experiments may be summarized through the following items:
\begin{enumerate}
	\item Extraction of Volumetric Bouligand-Minkowski descriptors from each image in the dataset;
	\item Computation of $\alpha$ coefficients of the approximating analytical function (functional data);
	\item Obtainment of the coefficients $\beta$ after the transform described in the Section \ref{sec:method};
	\item Use of coefficients $\alpha$ and $\beta$ as input to different classification methods;
	\item Comparison, in terms of classification performance, among the proposed approach and the direct use of volumetric Bouligand-Minkowski descriptors.
\end{enumerate}

The performance of the FDA transform is verified in direct approach (using $\alpha(u)$) and transformed approach (using $\beta(u)$). The basis used was B-spline. For the classification process we use classical methods from the literature \cite{DH00}, that is, Bayesian, K-Nearest Neighbor (KNN) and Linear Discriminant Analysis (LDA).

\section{Results}

The results are showed in graphs and tables which represent the different ways for the use of the FDA transform combined to Bouligand-Minkowksi descriptors in the datasets analyzed. Empirically, we found an optimal interval for the number of descriptors used, that is, between 60 and 100 for direct FDA coefficients and between 10 and 50 for transformed FDA fractal descriptors.

\begin{figure}[!hp]
					 \centering
           \mbox{\subfigure[]{\epsfig{figure=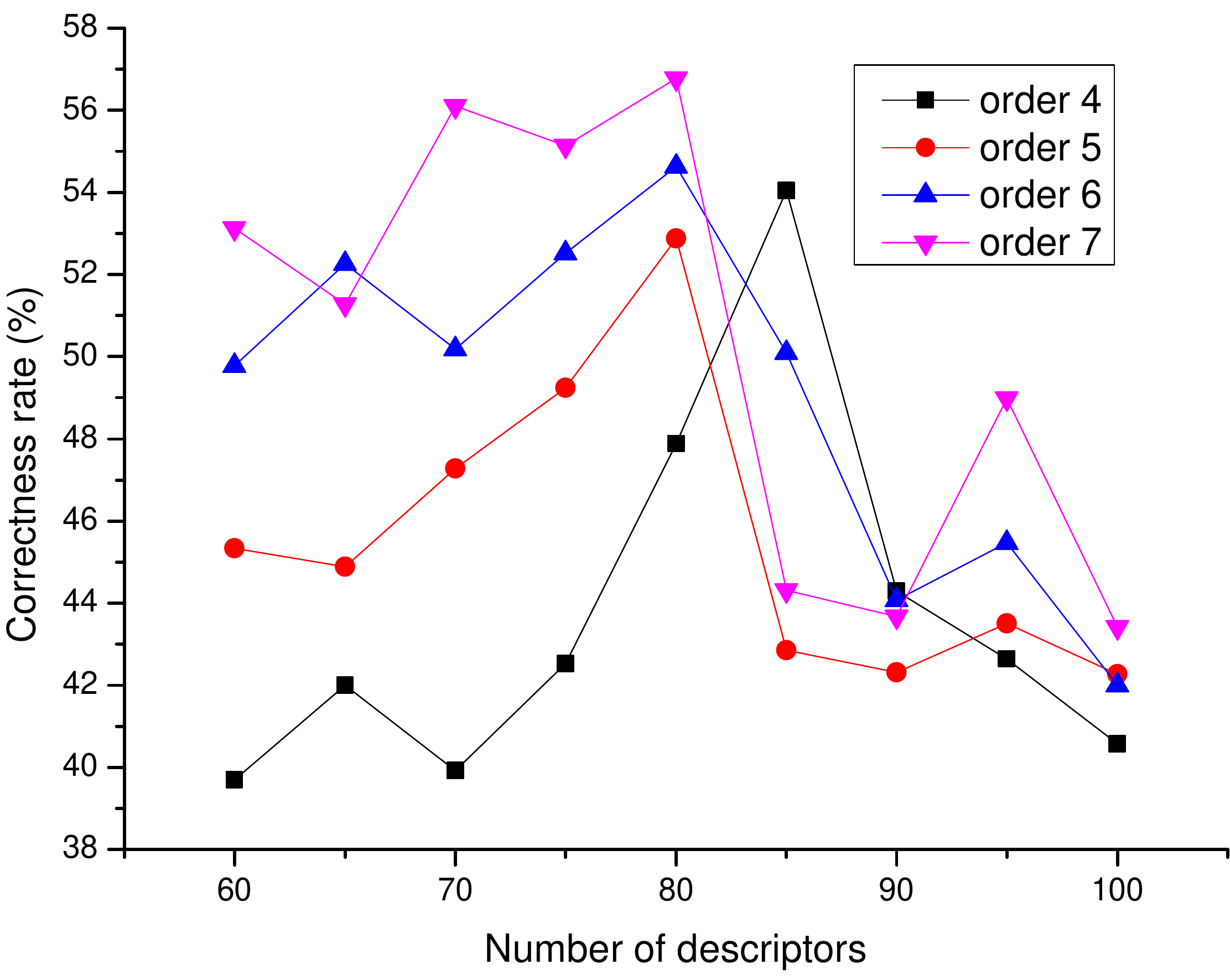,width=0.5\textwidth}}
           			 \subfigure[]{\epsfig{figure=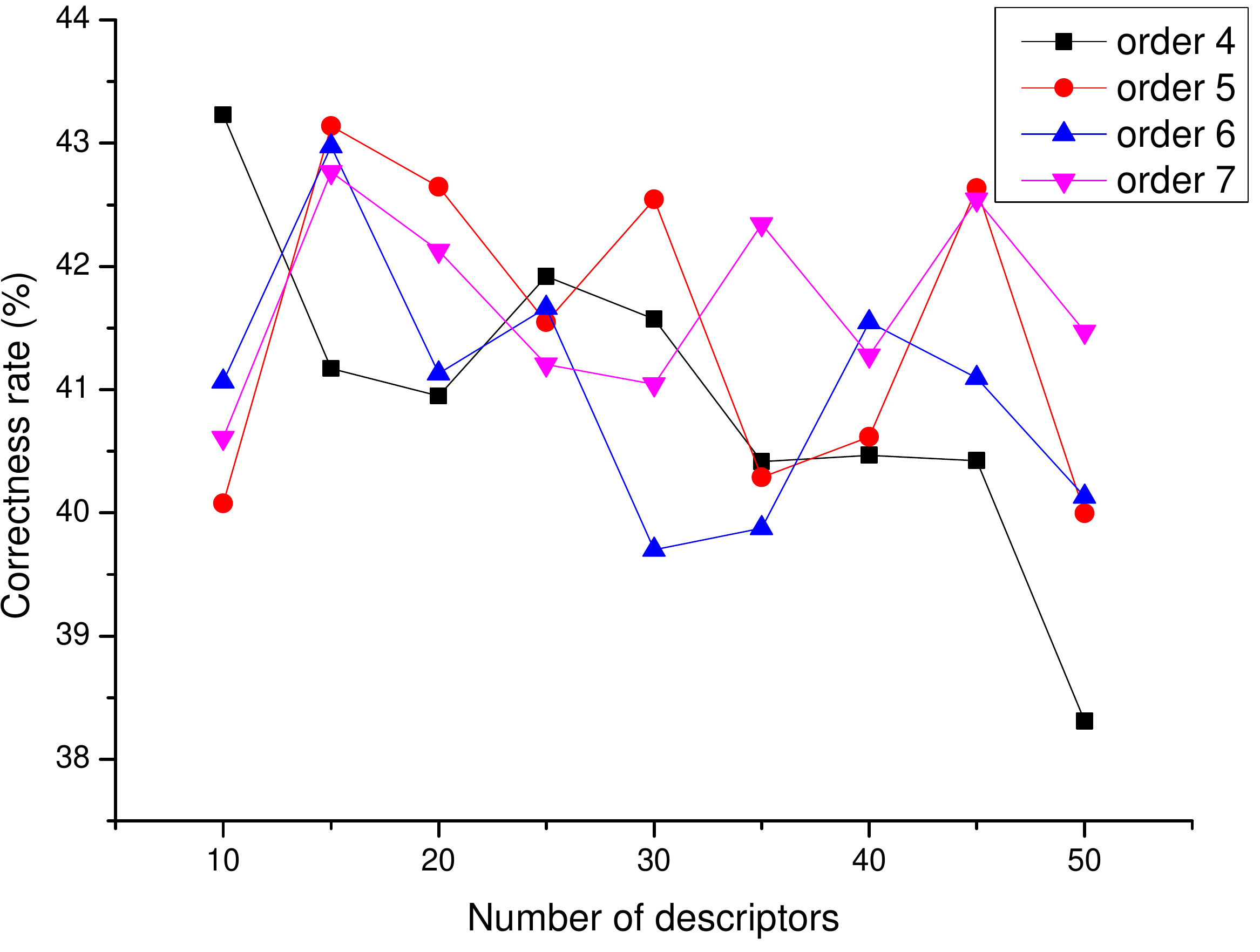,width=0.5\textwidth}}}
           \mbox{\subfigure[]{\epsfig{figure=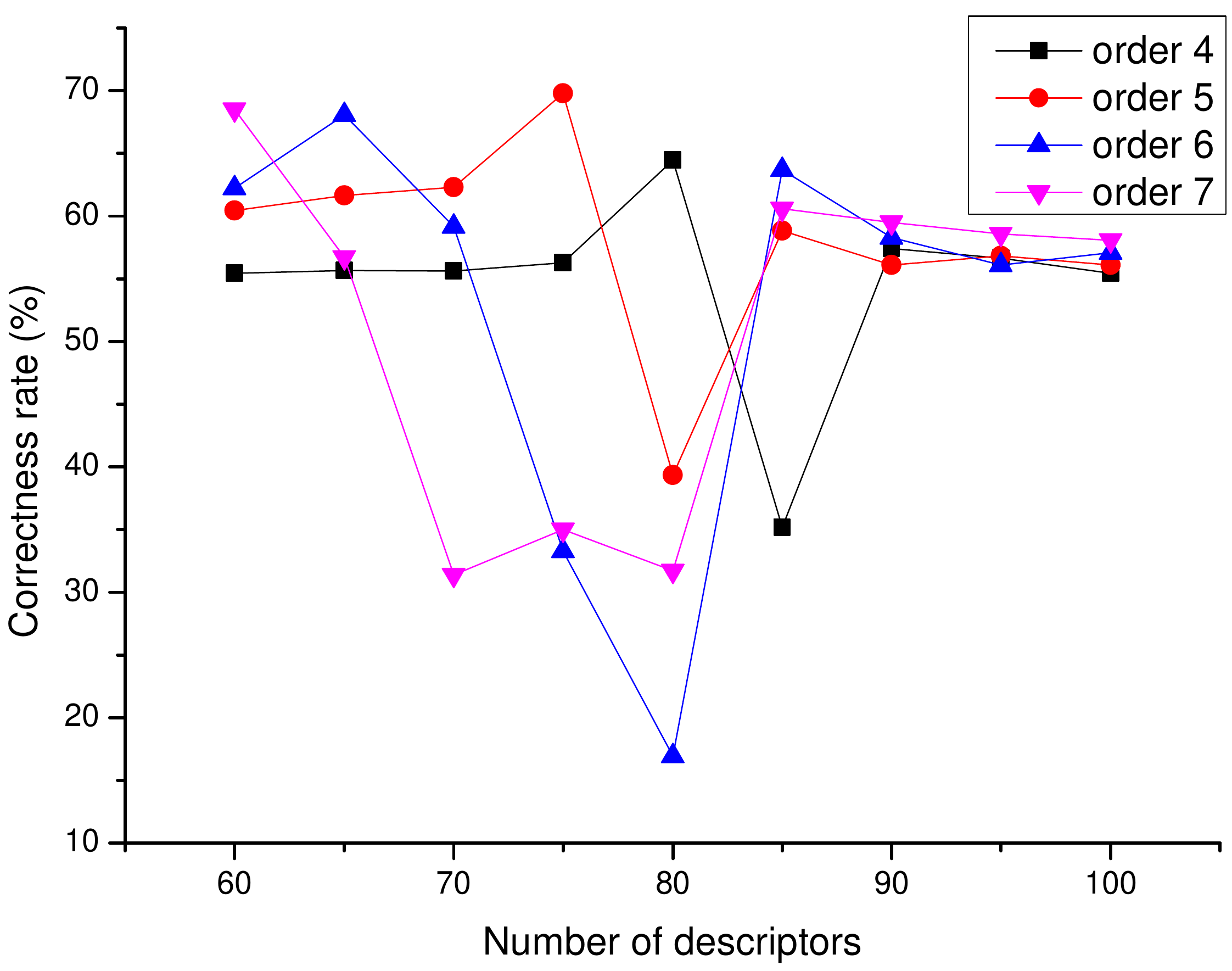,width=0.5\textwidth}}           			 
                 \subfigure[]{\epsfig{figure=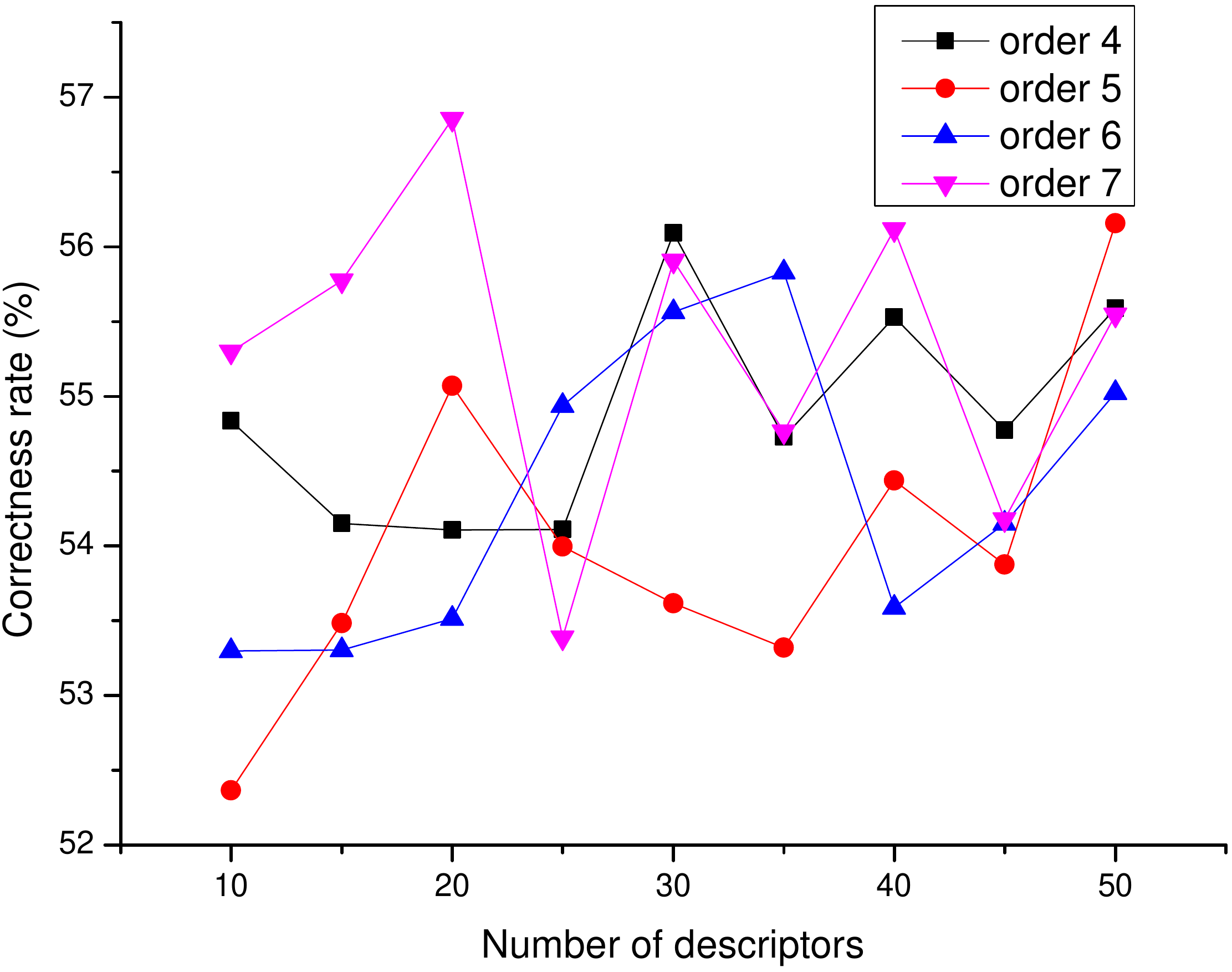,width=0.5\textwidth}}}
           \mbox{\subfigure[]{\epsfig{figure=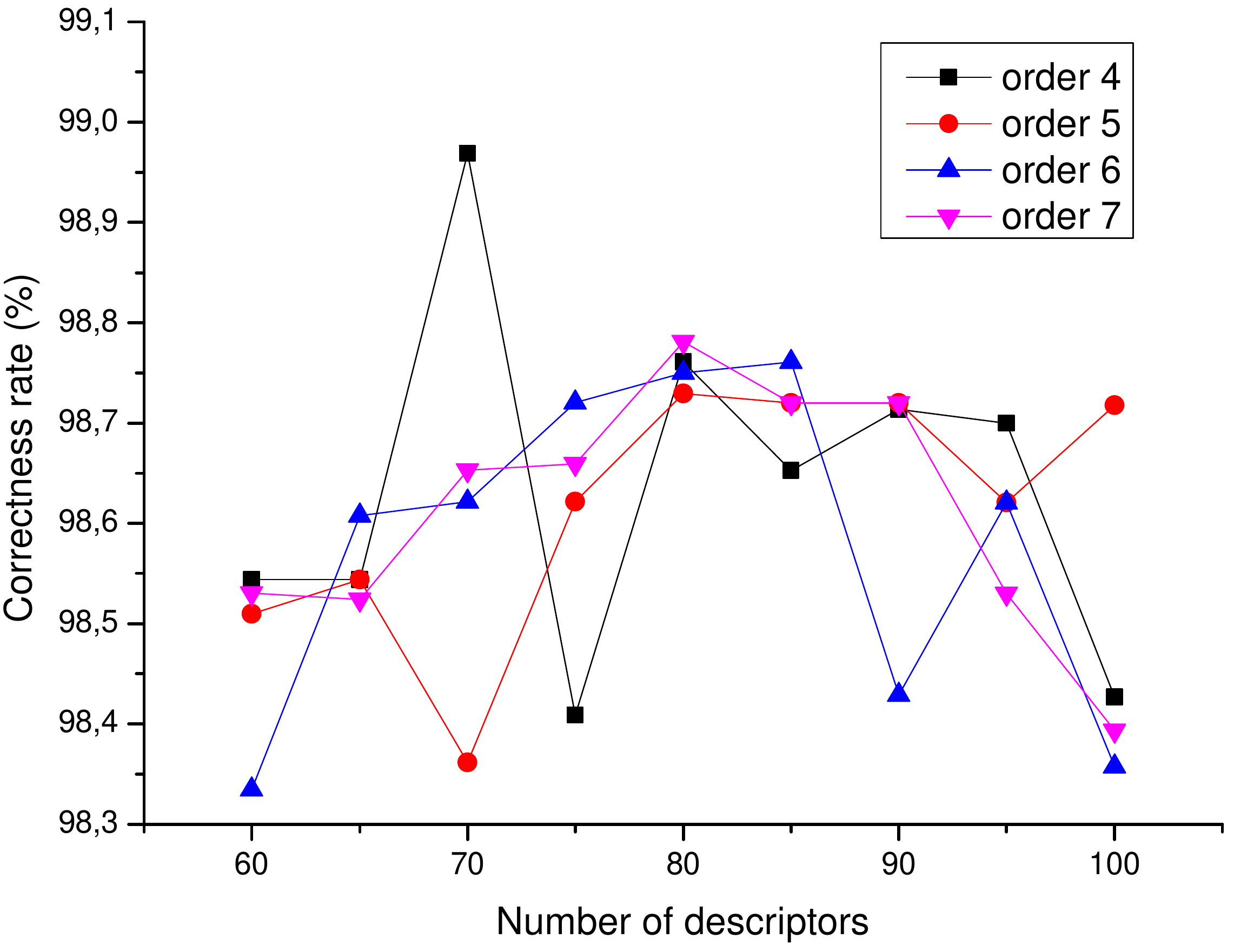,width=0.5\textwidth}}
           			 \subfigure[]{\epsfig{figure=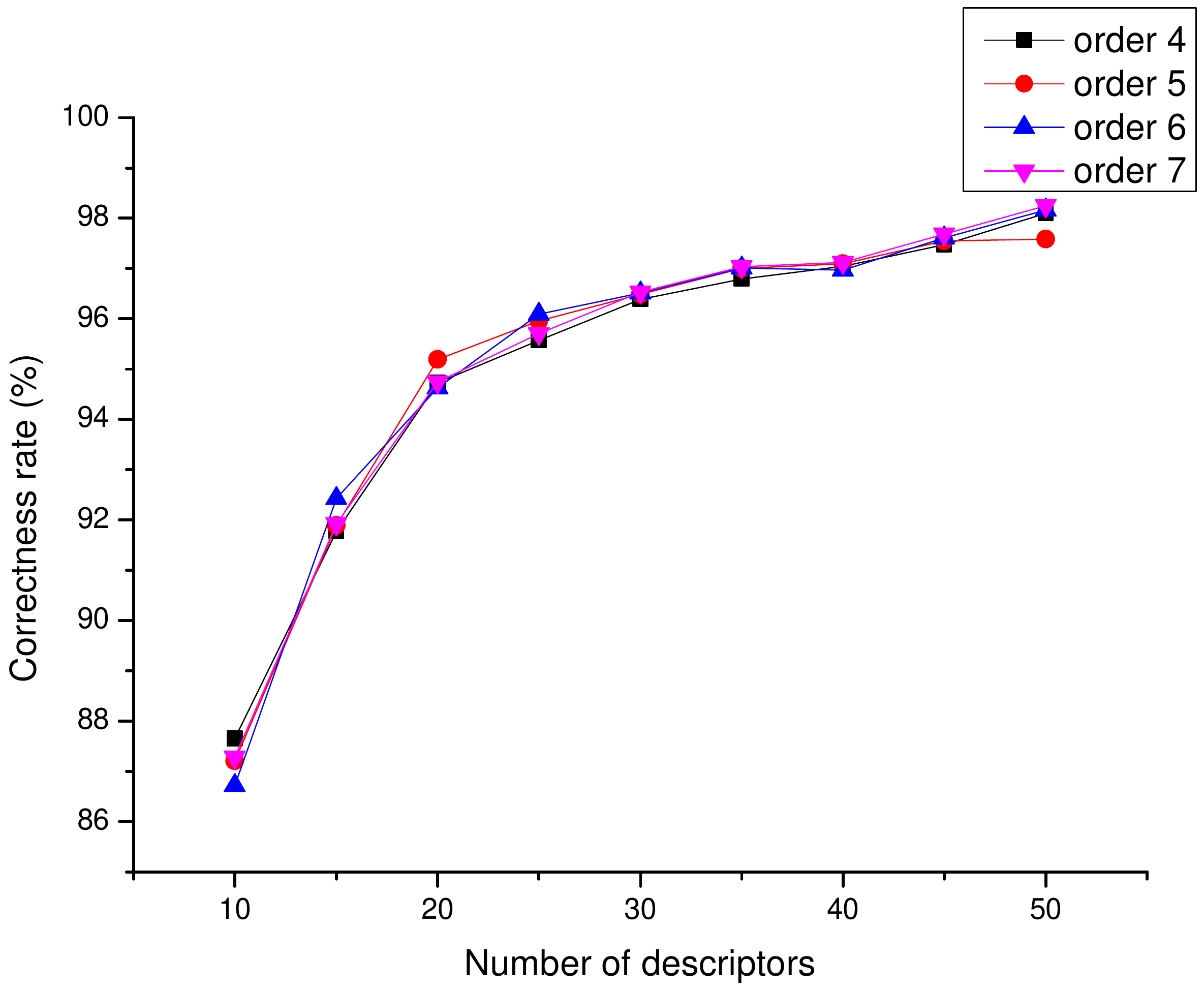,width=0.5\textwidth}}}           			            			 
           \caption{Correctness rate for Brodatz data set. At left, using normal FDA coefficients. At right, using transformed FDA descriptors. From above to below using Bayesian, KNN and LDA classifier.}
           \label{fig:resultbrod}                                  
   \end{figure}

Firstly, the Figure \ref{fig:resultbrod} shows the correctness rate for the use of FDA fractal descriptors in the classification of Brodatz data set. At left, we show the results for normal FDA. At right, for transformed FDA. From above to below, we use Bayesian, KNN and LDA classifier. Initially, we cannot notice any direct relation among number of descriptors, basis order and correctness rate. The exception occurs with the use of LDA with transformed FDA descriptors. In this case, it is clear that the correctness rate increases with the number of descriptors. In the most of cases, however, it is noticeable that higher order basis yield greater correctness. This is explained by the fact that those basis are capable of capture more details from the conventional VBFD descriptors. Relative to the number of descriptors, the graph shows that each specific combination of FDA descriptors and classifier provides a different pattern for the correctness rate results. This is also waited due to the fact that each classifier has a particular way of dealing with correlation and irregularity information.

The Figure \ref{tab:resultoutex} shows the correctness rate in Outex data set. The graphs are organized in the same way as in the Figure \ref{fig:resultbrod}. The observations from Brodatz data set are also valid in this case. Particularly, an interesting observation is that the aspect of the graph of each combination descriptor/classifier is similar in both data sets. The unique significant difference is the global correctness that is smaller in Outex, due to its greater difficulty level when compare to Brodatz data set. 

   \begin{figure}[!hp] 
					 \centering
           \mbox{\subfigure[]{\epsfig{figure=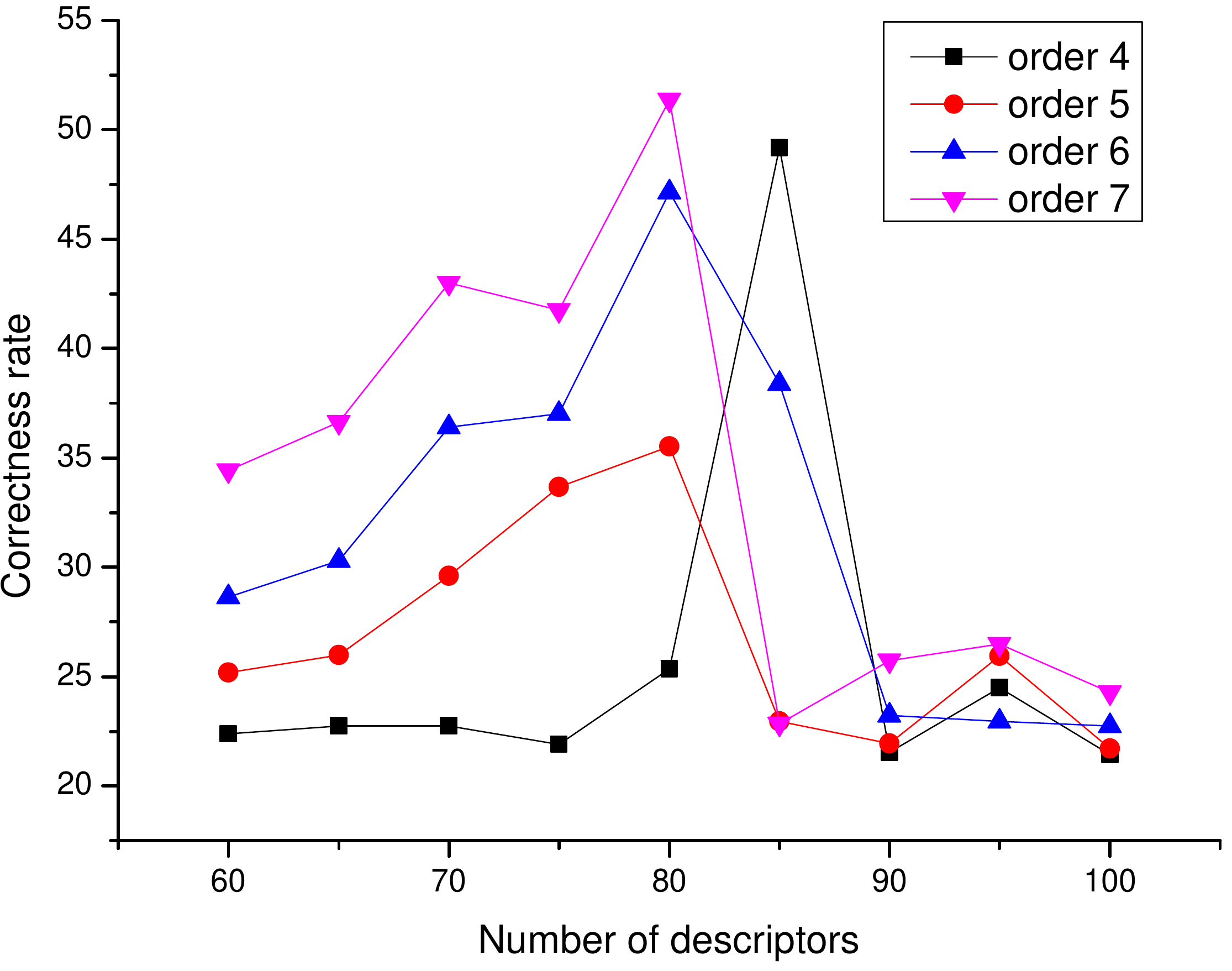,width=0.5\textwidth}}
           			 \subfigure[]{\epsfig{figure=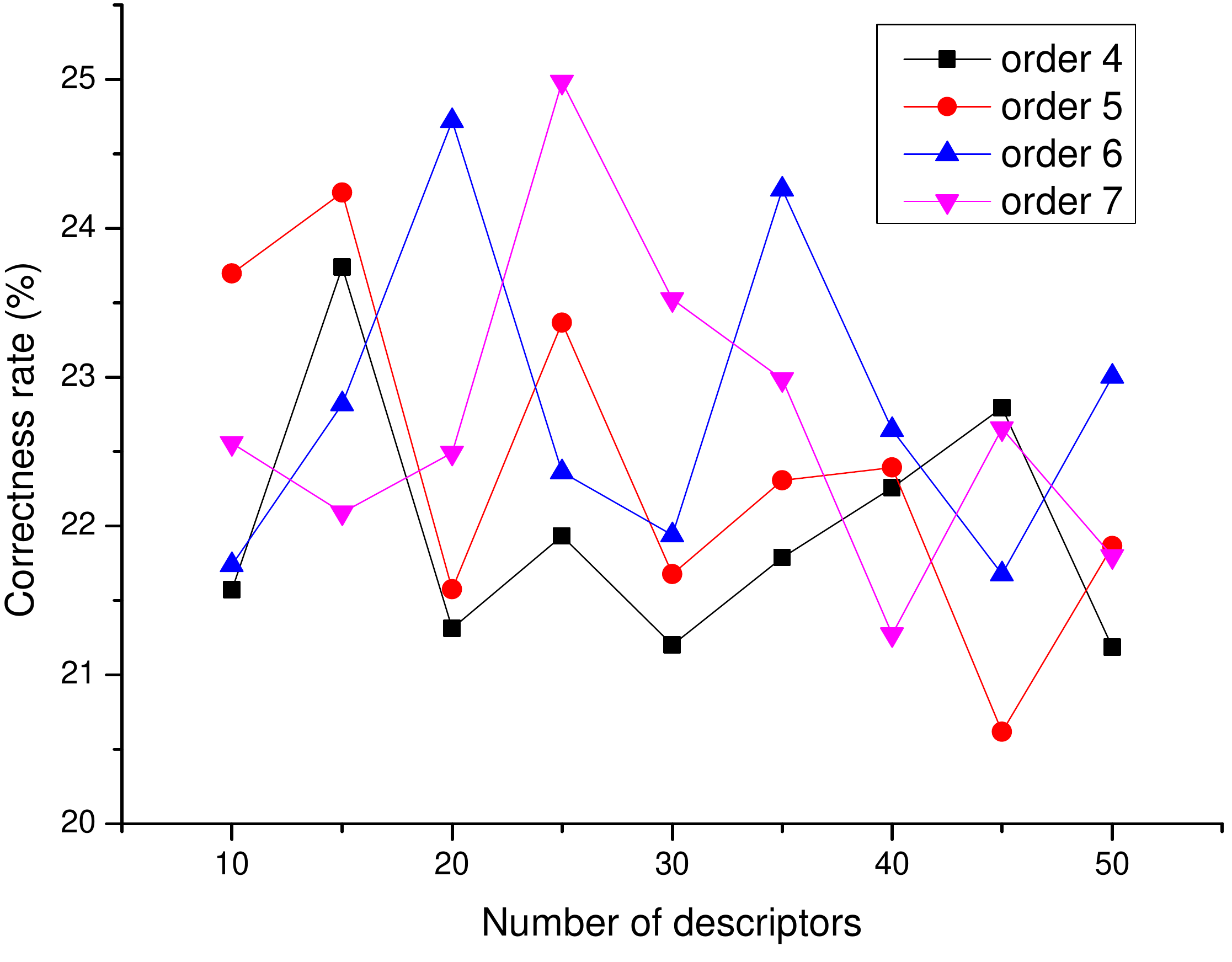,width=0.5\textwidth}}}
           \mbox{\subfigure[]{\epsfig{figure=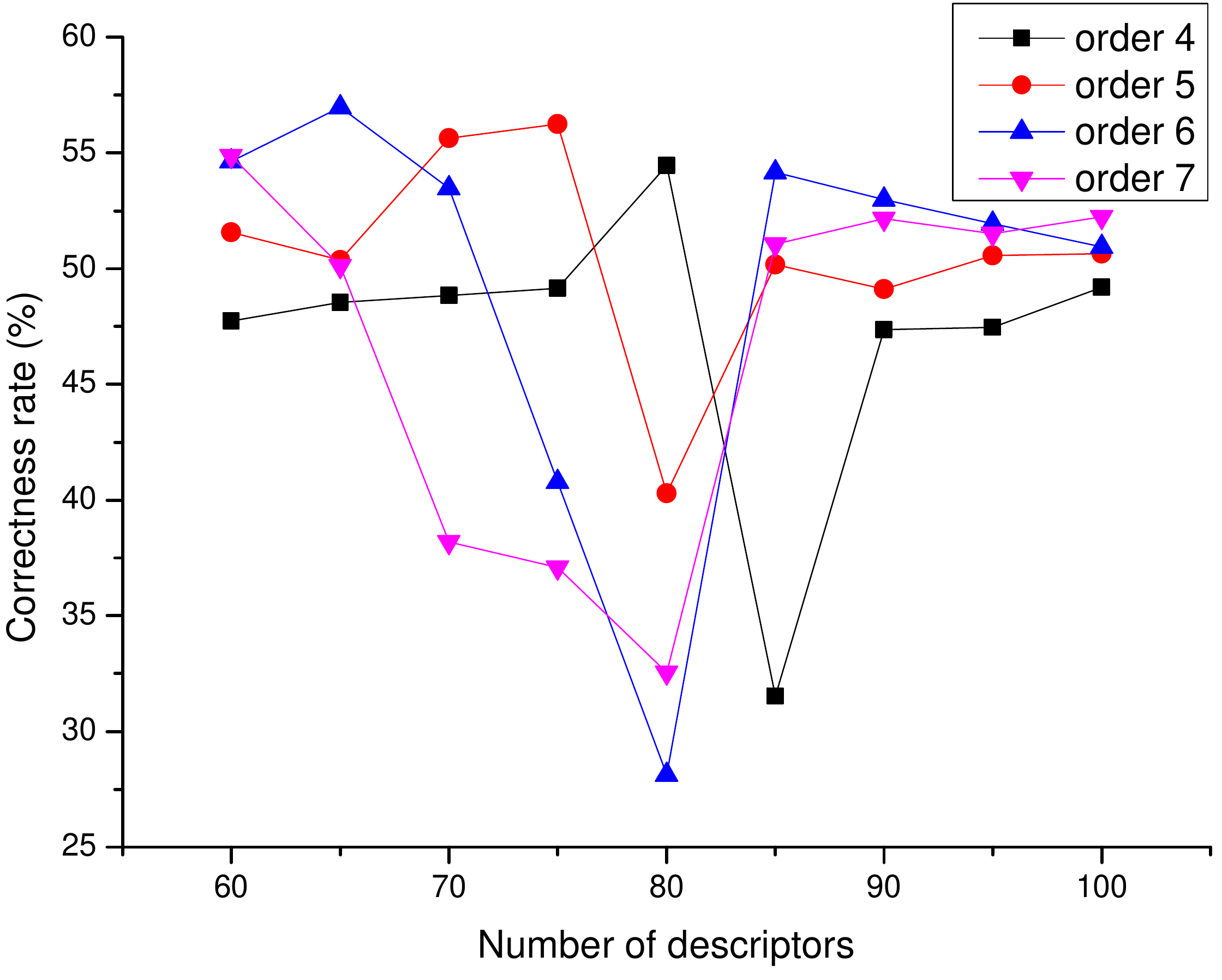,width=0.5\textwidth}}           			 
                 \subfigure[]{\epsfig{figure=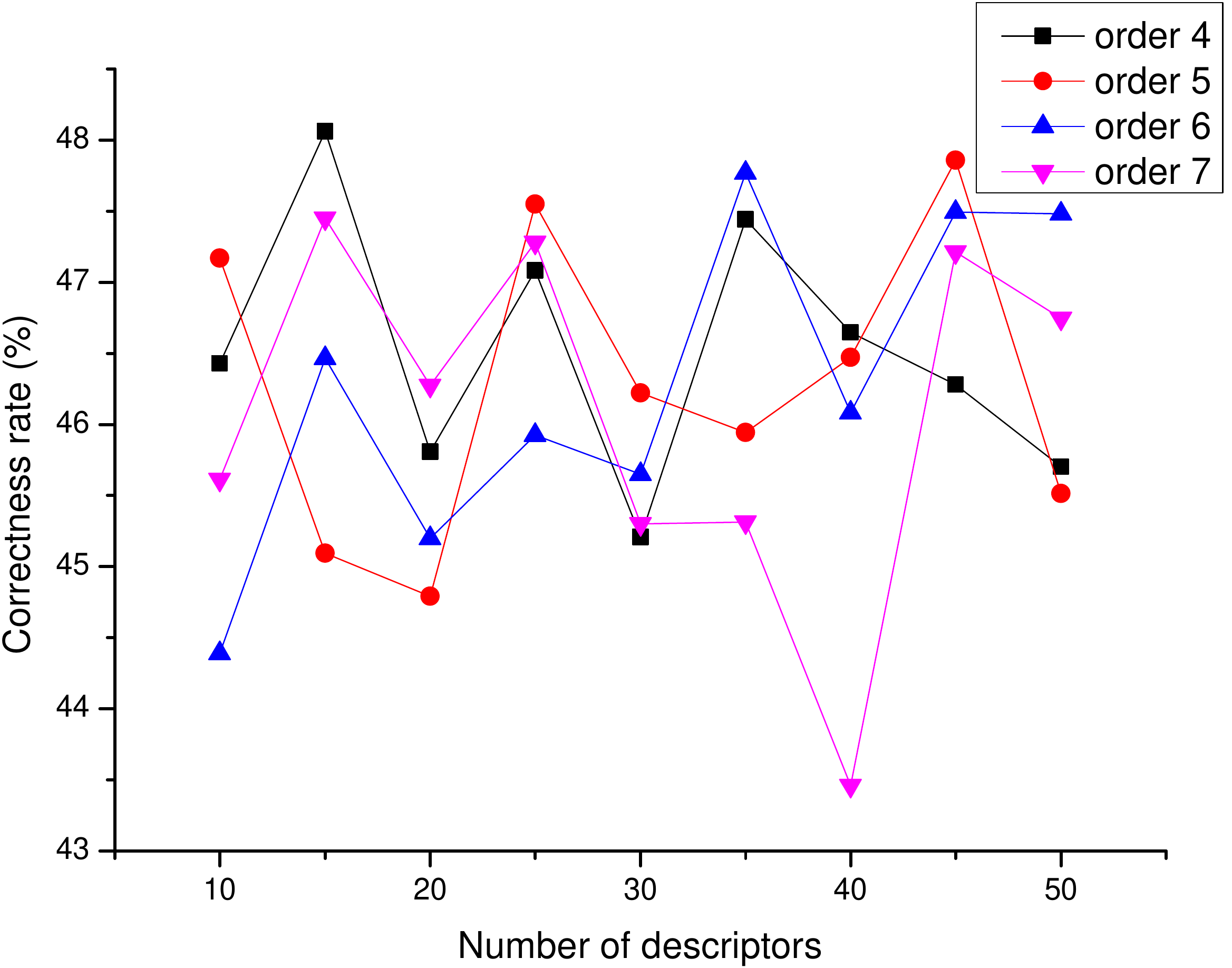,width=0.5\textwidth}}}
           \mbox{\subfigure[]{\epsfig{figure=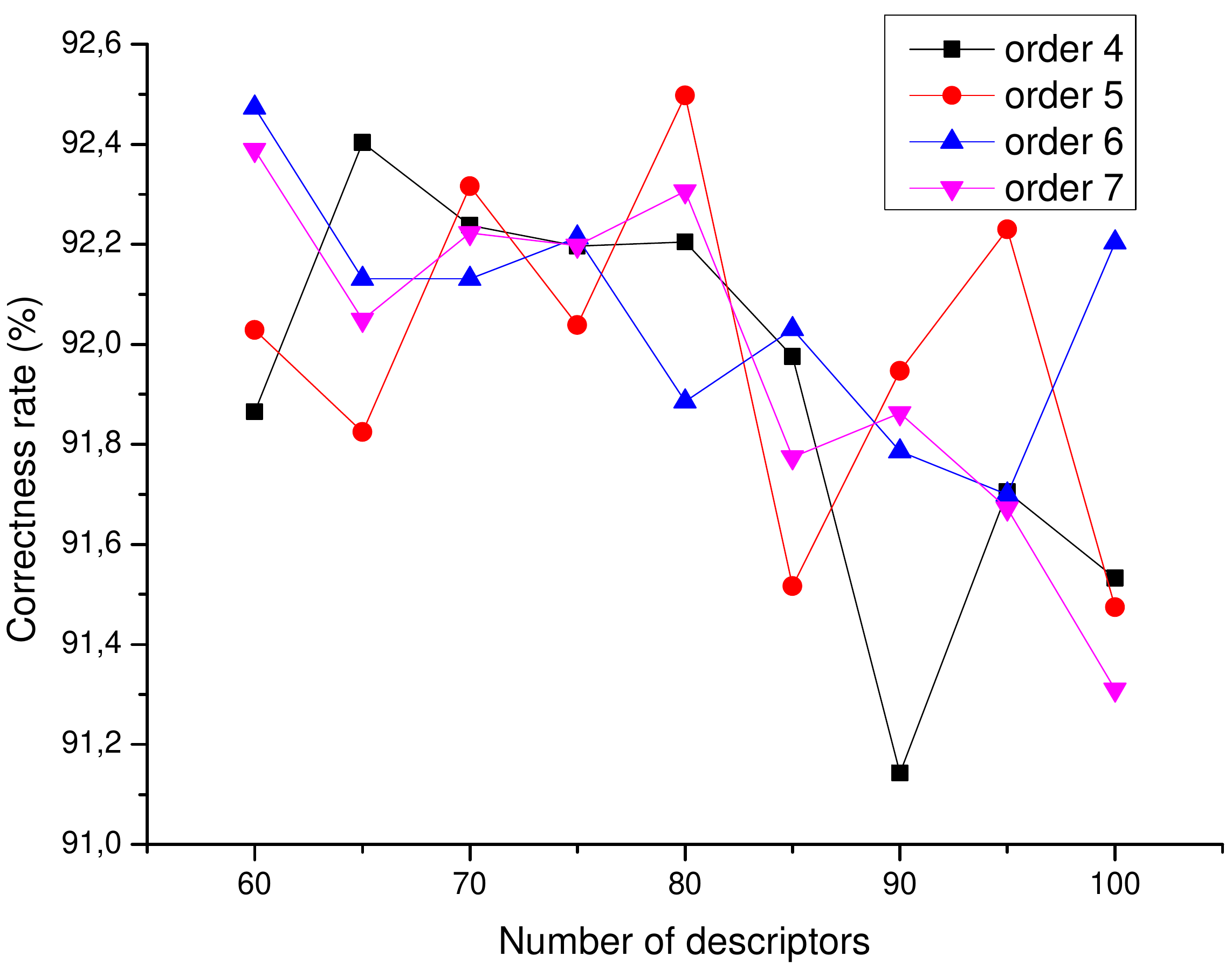,width=0.5\textwidth}}
           			 \subfigure[]{\epsfig{figure=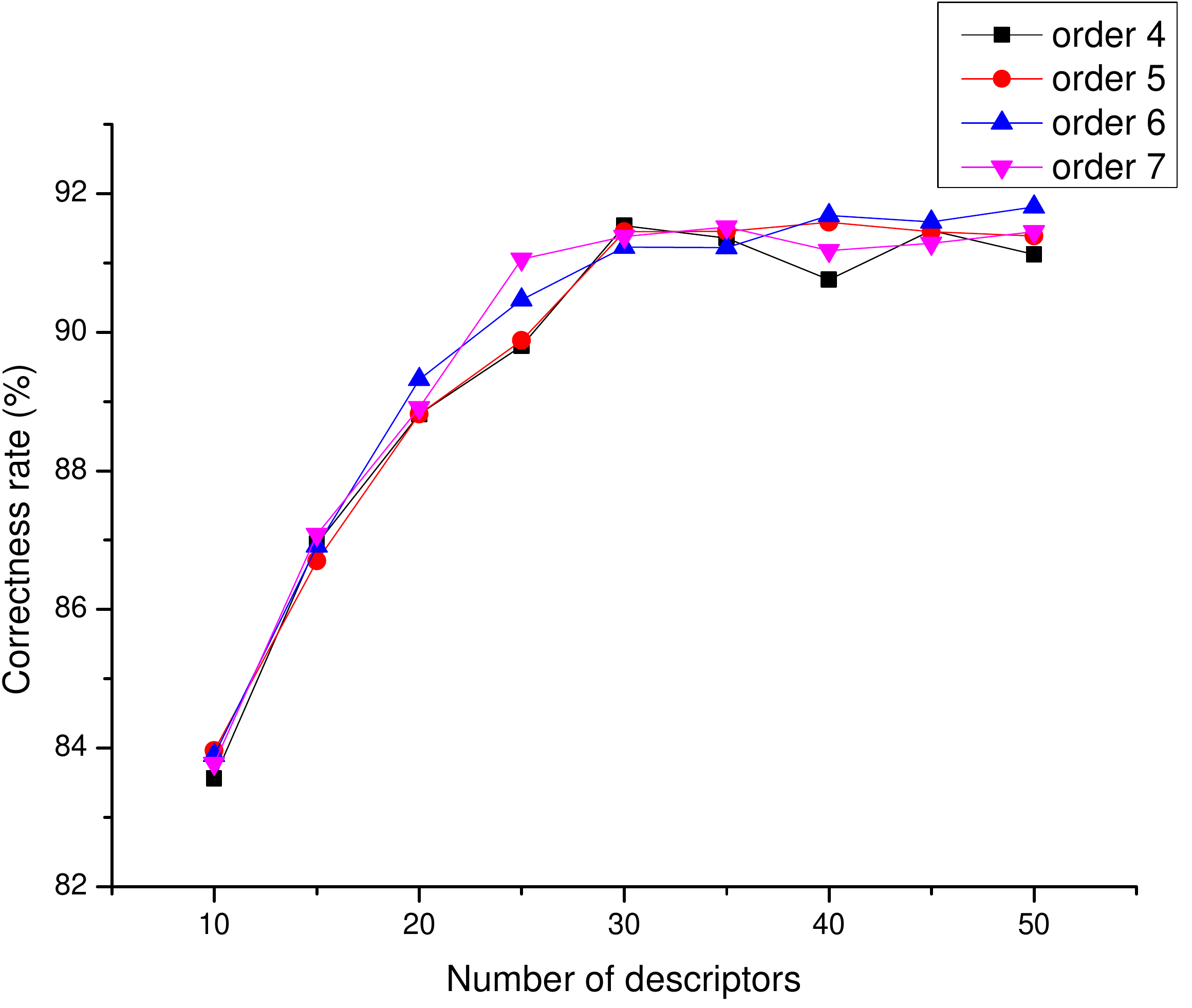,width=0.5\textwidth}}}           			            			 
           \caption{Correctness rate for Outex data set. At left, using normal FDA coefficients. At right, using transformed FDA descriptors. From above to below using Bayesian, KNN and LDA classifier.}
           \label{fig:resultoutex}                                  
   \end{figure}

Now we show the best results achieved by each combination of descriptors and classifiers and the number of used descriptors. In the Table \ref{tab:resultbrod} we show the correctness rate for Brodatz data set. We observe that even using a reduced set of descriptors, FDA achieved results sensibly more precise than classical VB fractal descriptors. This advantage is more notable in KNN and Bayesian classifier. In Bayesian, FDA presented an advantage of 42\% while in KNN this advantage was 27\%. Another important observation from the table is that in this specific application the use of normal FDA coefficients demonstrated to be the better solution. This is very encouraging since this FDA approach is computationally simple and allows an easy statistical interpretation of the analyzed data.

\begin{table*}[!htpb]
	\centering
	\scriptsize
		\begin{tabular}{ccccc}
\hline		
Descriptors & Classifier & Number of descriptors & Correctness rate (\%)\\
\hline
Original & Bayesian & 86 & 40.0 $\pm$ 0.2\\
Normal FDA & Bayesian & 80 & 56.8 $\pm$ 0.1\\
Transformed FDA & Bayesian & 10 & 43.2 $\pm$ 0.1\\
\hline
Original & KNN & 86 & 54.8 $\pm$ 0.1\\
Normal FDA & KNN & 75 & 69.8 $\pm$ 0.2\\
Transformed FDA & KNN & 20 & 56.8 $\pm$ 0.1\\
\hline
Original & LDA & 86 & 98.6 $\pm$ 0.1\\
Normal FDA & LDA & 70 & 99.0 $\pm$ 0.1\\
Transformed FDA & LDA & 50 & 98.2 $\pm$ 0.0\\
\hline
		\end{tabular}
	\caption{Correctness rate for the use of fractal descriptors enhanced by FDA normal and transformed coefficients in Brodatz dataset.}
	\label{tab:resultbrod}
\end{table*}

Concluding, we present the results of FDA descriptors in OuTex dataset. Again, the performance of FDA descriptors is very good. A highlight must be given to the Bayesian result. FDA provided a correctness rate 123\% greater than classical fractal descriptors.

\begin{table*}[!htpb]
	\centering
	\scriptsize
		\begin{tabular}{ccccc}
\hline		
Descriptors & Classifier & Number of descriptors & Correctness rate (\%)\\
\hline
Original & Bayesian & 86 & 23.0 $\pm$ 0.1\\
Normal FDA & Bayesian & 80 & 51.4 $\pm$ 0.2\\
Transformed FDA & Bayesian & 25 & 24.9 $\pm$ 0.1\\
\hline
Original & KNN & 86 & 47.3 $\pm$ 0.2\\
Normal FDA & KNN & 65 & 57.0 $\pm$ 0.1\\
Transformed FDA & KNN & 15 & 48.1 $\pm$ 0.1\\
\hline
Original & LDA & 86 & 92.0 $\pm$ 0.0\\
Normal FDA & LDA & 80 & 92.5 $\pm$ 0.1\\
Transformed FDA & LDA & 50 & 91.8 $\pm$ 0.2\\
\hline
		\end{tabular}
	\caption{Correctness rate for the use of fractal descriptors enhanced by FDA normal and transformed coefficients in Outex dataset.}
	\label{tab:resultoutex}
\end{table*}

From the previous results, we observe that we cannot extract an exact relation between the number of FDA basis (and, consequently, descriptors), the order of used basis and the correctness rate results. However, analyzing without excessive severity, we observe that generally the use of higher order basis increases the classification performance. Nevertheless, it is always important to verify the combinations for each different application.

Analyzing more globally the results, we observe initially that the FDA transform provides a significant increasing in the performance of volumetric Bouligand-Minkowski descriptors, mainly when we used KNN and Bayesian classifier. This fact attests that the FDA transform is capable of extract relevant features from the set of descriptors, allowing for the classifiers to provide a more precise classification result. As discussed in the Section \ref{sec:method}, the good performance of the FDA transform was expected due to the smooth analytical and irregularly spaced nature of Bouligand-Minkowski descriptors. The smaller efficiency in LDA classifier is easily explained by the fact that one step in the LDA method involves a correlation space transform (Principal Component Analysis). So, features extracted by the FDA transform do not necessarily to have the same correlation properties as the original descriptors and this fact prejudices the performance of the whole classification process.

\section{Conclusions}

This work proposed and analyzed the use of the FDA transform aiming at enhancing the performance of volumetric Bouligand-Minkowski fractal descriptors, applied to texture classification. The transform consists in the use of coefficients from the functional data representation replacing the original descriptors.

Results demonstrated that the FDA transform increased significantly the accuracy of classification process, mainly when using Bayesian and KNN classifiers. Results confirmed what is expected from the theory, once FDA is a powerful statistical tool for the representation of smooth analytical data, like fractal descriptors. The FDA transform extracts important features and patterns from the original descriptors set yielding a better classification performance.

Results suggest strongly that FDA must be considered as an auxiliary tool for other methods shown in the literature for obtaining fractal descriptors or even other techniques in texture analysis that generate a set of values which may be handled as an analytical function.

\section{Acknowledgements}
\label{sec:Acknowledgements}
Odemir M. Bruno gratefully acknowledges the financial support of CNPq (National Council for Scientific and Technological Development, Brazil) (Grant \#308449/2010-0 and \#473893/2010-0) and FAPESP (The State of S\~ao Paulo Research Foundation) (Grant \# 2011/01523-1). Jo\~ao B. Florindo is grateful to CNPq (National Council for Scientific and Technological Development, Brazil) for his doctorate grant.

\newpage


\end{document}